\documentclass[10pt,twocolumn,letterpaper]{article}

\usepackage{cvpr}              

\usepackage{algorithm}
\usepackage{algorithmic}
\usepackage{amsmath}
\usepackage{multirow}
\usepackage{makecell}
\usepackage{subcaption}
\usepackage{array}
\usepackage{xcolor}
\usepackage{balance}
\usepackage{booktabs}       
\usepackage{graphicx}       
\usepackage{array}          
\usepackage[x11names]{xcolor}  
\usepackage{colortbl}       
\definecolor{cvprblue}{rgb}{0.21,0.49,0.74}
\usepackage[pagebackref,breaklinks,colorlinks,allcolors=cvprblue]{hyperref}

\def\paperID{3127} 
\def\confName{CVPR}
\def\confYear{2026}

\title{Test-time Ego-Exo-centric Adaptation for Action Anticipation via Multi-Label\\Prototype Growing and Dual-Clue Consistency}

\author{Zhaofeng Shi, Heqian Qiu\footnotemark[1], Lanxiao Wang\footnotemark[1], Qingbo Wu, Fanman Meng, Lili Pan, Hongliang Li\footnotemark[1]\\
University of Electronic Science and Technology of China, Chengdu, China\\
{\tt\small \{zfshi\}@std.uestc.edu.cn \quad \{hqqiu,lanxiaowang,qbwu,fmmeng,lilipan,hlli\}@uestc.edu.cn} 
}

\begin{document}
\maketitle

\footnotetext[1]{Corresponding authors.}

\begin{abstract}
Efficient adaptation between Egocentric (Ego) and Exocentric (Exo) views is crucial for applications such as human-robot cooperation. However, the success of most existing Ego-Exo adaptation methods relies heavily on target-view data for training, thereby increasing computational and data collection costs. In this paper, we make the first exploration of a Test-time Ego-Exo Adaptation for Action Anticipation (TE$^{2}$A$^{3}$) task, which aims to adjust the source-view-trained model online during test time to anticipate target-view actions. It is challenging for existing Test-Time Adaptation (TTA) methods to address this task due to the multi-action candidates and significant temporal-spatial inter-view gap. Hence, we propose a novel Dual-Clue enhanced Prototype Growing Network (DCPGN), which accumulates multi-label knowledge and integrates cross-modality clues for effective test-time Ego-Exo adaptation and action anticipation. Specifically, we propose a Multi-Label Prototype Growing Module (ML-PGM) to balance multiple positive classes via multi-label assignment and confidence-based reweighting for class-wise memory banks, which are updated by an entropy priority queue strategy. Then, the Dual-Clue Consistency Module (DCCM) introduces a lightweight narrator to generate textual clues indicating action progressions, which complement the visual clues containing various objects. Moreover, we constrain the inferred textual and visual logits to construct dual-clue consistency for temporally and spatially bridging Ego and Exo views. Extensive experiments on the newly proposed EgoMe-anti and the existing EgoExoLearn benchmarks show the effectiveness of our method, which outperforms related state-of-the-art methods by a large margin. Code is available at \href{https://github.com/ZhaofengSHI/DCPGN}{https://github.com/ZhaofengSHI/DCPGN}.

\vspace{-3mm}

\end{abstract}

\section{Introduction}
By virtue of mirror neurons \cite{rizzolatti2004mirror,iacoboni2009imitation}, humans can seamlessly switch between exocentric (Exo) and egocentric (Ego) views and map out the upcoming activities in the future. For example, a person may observe a cooking process from an Exo view, then naturally understand and take over the remaining operations in Ego perspective. This amazing capability is also fundamental in computer vision systems with a wide range of applications, such as human-robot cooperation \cite{kang2023video,shi2024cross,kozamernik2023visual} and embodied AI \cite{duan2022survey,wang2024embodiedscan,zhang2020language}.

Despite the success in studies on action recognition \cite{wang2023ego,jhuang2013towards,wang2021interactive,sun2022human} or anticipation \cite{gammulle2019predicting,gong2022future,mittal2024can} in a single view, the models lapse when applied to the other perspective due to completely different camera recording angles and styles. Therefore, an increasing number of works \cite{huang2024egoexolearn,grauman2024ego,qiu2025egome,li2024egoexo,kwon2021h2o} have attempted to replicate the aforementioned human instinct of Ego-Exo adaptation. In detail, some works follow a pretrain-finetune scheme \cite{xue2023learning,luo2025viewpoint,ohkawa2023exo2egodvc,reilly2025my}, which first performs source-view training or cross-view representation learning and then uses labeled data for finetuning. For instance, Exo2EgoDVC \cite{ohkawa2023exo2egodvc} leverages large-scale Exo videos to augment Ego dense video captioning. AE2 \cite{xue2023learning} learns view-invariant representations via unpaired videos to facilitate multiple downstream tasks. Although having achieved desirable performance, they require labour-intensive target-view labeling for supervision. To alleviate this limitation, some works adopt the idea of unsupervised domain adaptation (UDA) \cite{ganin2015unsupervised,yang2023tvt,shi2025unsupervised,tzeng2014deep}, like Sync \cite{quattrocchi2024synchronization} using the labeled Exo data and unlabeled synchronized Ego-Exo pairs for temporal action segmentation in the Ego view. GCEAN \cite{shi2025unsupervised} transfers knowledge from the labeled source view to the unlabeled target view during training for procedural activity understanding. However, these methods still require access to unlabeled target-view data for training, creating a bottleneck in their employment due to the additional computational and data collection costs. Consequently, we elicit a question: \textit{Can we directly adapt the model to anticipate the target-view actions without re-training requirements?}

\begin{figure}[!t]
\centering
\includegraphics[width=0.96\linewidth]{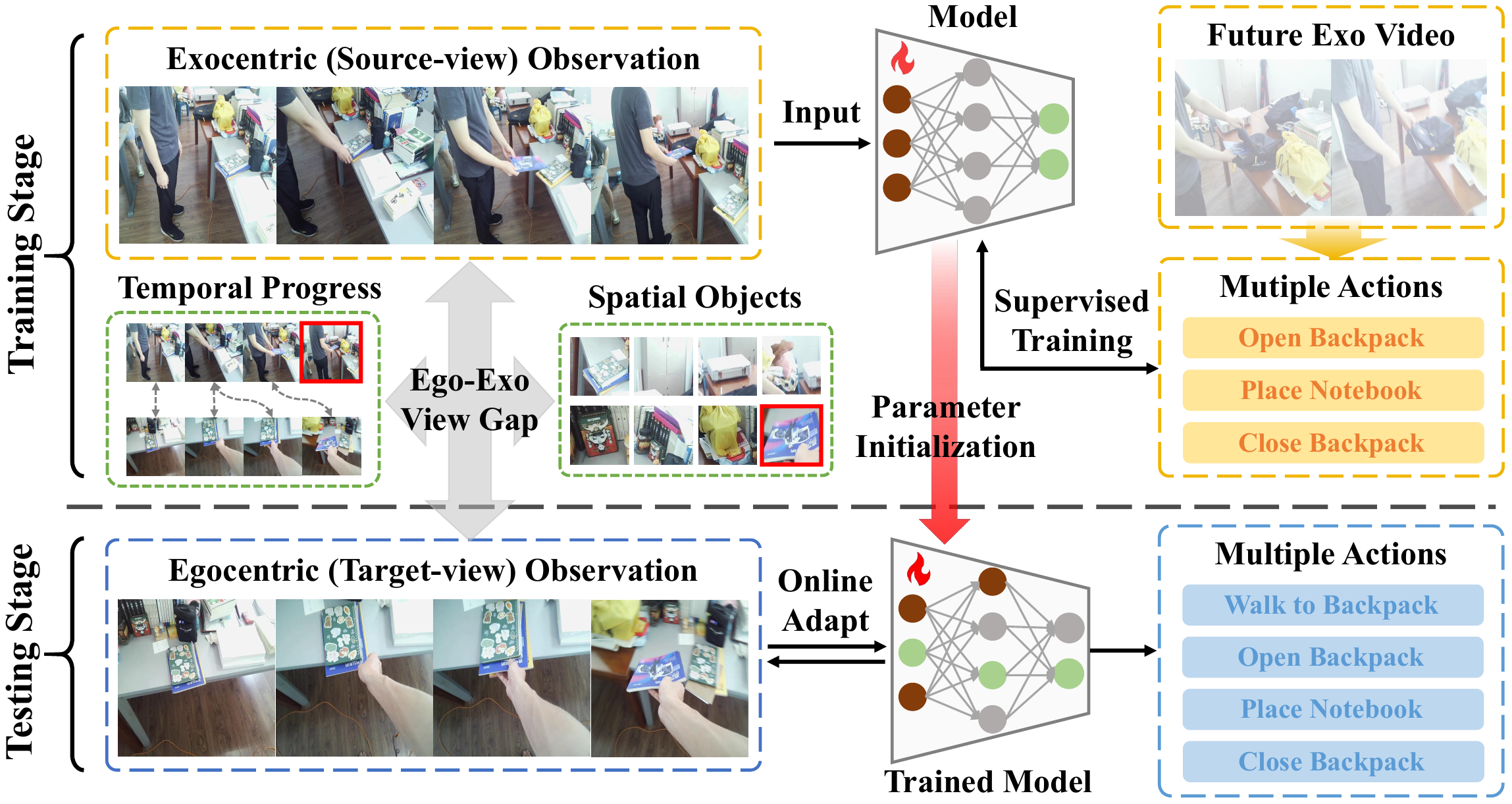}
\caption{Schematic of the newly proposed TE$^{2}$A$^{3}$ task.}
\label{fig:1}
\vspace{-4mm}
\end{figure}

In this paper, we formulate this question as a Test-Time Adaptation (TTA) \cite{wang2020tent,shu2022test,zhang2022tip,farina2024frustratingly,feng2023diverse,wang2024less} problem, and make the first exploration of a Test-time Ego-Exo Adaptation for Action Anticipation (TE$^{2}$A$^{3}$) task as shown in Fig. \ref{fig:1}. This task aims to adjust the source-view-trained model online during test time to anticipate the upcoming actions in the target view. This task is challenging due to its multi-action candidates and significant Ego-Exo spatial-temporal gap. Specifically, on the one hand, most TTA methods \cite{wang2020tent,shu2022test,lin2023video} use entropy-based or cache-based strategies to emphasize a single class with the highest confidence. However, it leads to sub-optimal performance in the proposed task because each event in the real-world scenarios typically comprises multiple atomic actions, which need to be anticipated simultaneously. On the other hand, despite the success of current methods \cite{wu2025multi, karmanov2024efficient} in image-level cross-domain TTA, they struggle with the adaptation between Ego and Exo videos, which exhibit a more remarkable inter-view gap in spatial (i.e., inconsistent layout and interferential objects) and temporal (i.e., asynchronous progress of actions) dimensions shown in green dashed boxes in Fig. \ref{fig:1}.

To tackle the above problems, we propose a novel Dual-Clue enhanced Prototype Growing Network (DCPGN), which progressively accumulates multi-label knowledge and integrates clues from different modalities for effective test-time Ego-Exo adaptation and action anticipation. In detail, to prevent the model from being biased to the most confident class while overlooking others, we propose a Multi-Label Prototype Growing Module (ML-PGM). It assigns multiple pseudo labels to the extracted representations and performs confidence-based reweighting to balance multiple positive classes. Meanwhile, an entropy priority queue strategy is also applied to update reliable representations for class-wise memory banks. In addition, we leverage an on-the-shelf CLIP model and propose a Dual-Clue Consistency Module (DCCM). Specifically, we introduce a lightweight narrator to generate descriptions as textual clues, which naturally serve as indicators for temporal action progressions to complement visual clues containing various spatial objects. Then, we constrain their inferred logits to construct the dual-clue consistency for explicitly bridging the Ego and Exo videos in temporal and spatial dimensions. For a comprehensive evaluation, we not only evaluate our method on the existing \textit{EgoExoLearn} benchmark, but also construct a brand-new \textit{EgoMe-anti} benchmark based on the EgoMe \cite{qiu2025egome} dataset, which captures paired videos from the perspectives of observer and follower in diverse real-life scenarios.

The major contributions can be concluded as follows:

\begin{itemize}

\item To the best of our knowledge, it is the first exploration of the Test-time Ego-Exo Adaptation for Action Anticipation (TE$^{2}$A$^{3}$) task. Moreover, we propose a Dual-Clue enhanced Prototype Growing Network (DCPGN) to progressively accumulate multi-label knowledge and integrate clues from different modalities for effective test-time Ego-Exo adaptation and action anticipation.

\item We develop a Multi-Label Prototype Growing Module (ML-PGM) for balancing multiple positive classes via multi-label assignment and confidence-based reweighting under an entropy priority queue strategy. And the Dual-Clue Consistency Module (DCCM) introduces the textual and visual clues to construct dual-clue consistency to explicitly bridge the Ego-Exo temporal-spatial gap.

\item We construct a brand-new \textit{EgoMe-anti} benchmark, and extensive experiments on the \textit{EgoExoLearn} and \textit{EgoMe-anti} benchmarks show that our method outperforms the related state-of-the-art methods by a large margin.

\end{itemize}

\section{Related Work}

\begin{figure*}[!t]
\centering
\includegraphics[width=0.92\linewidth]{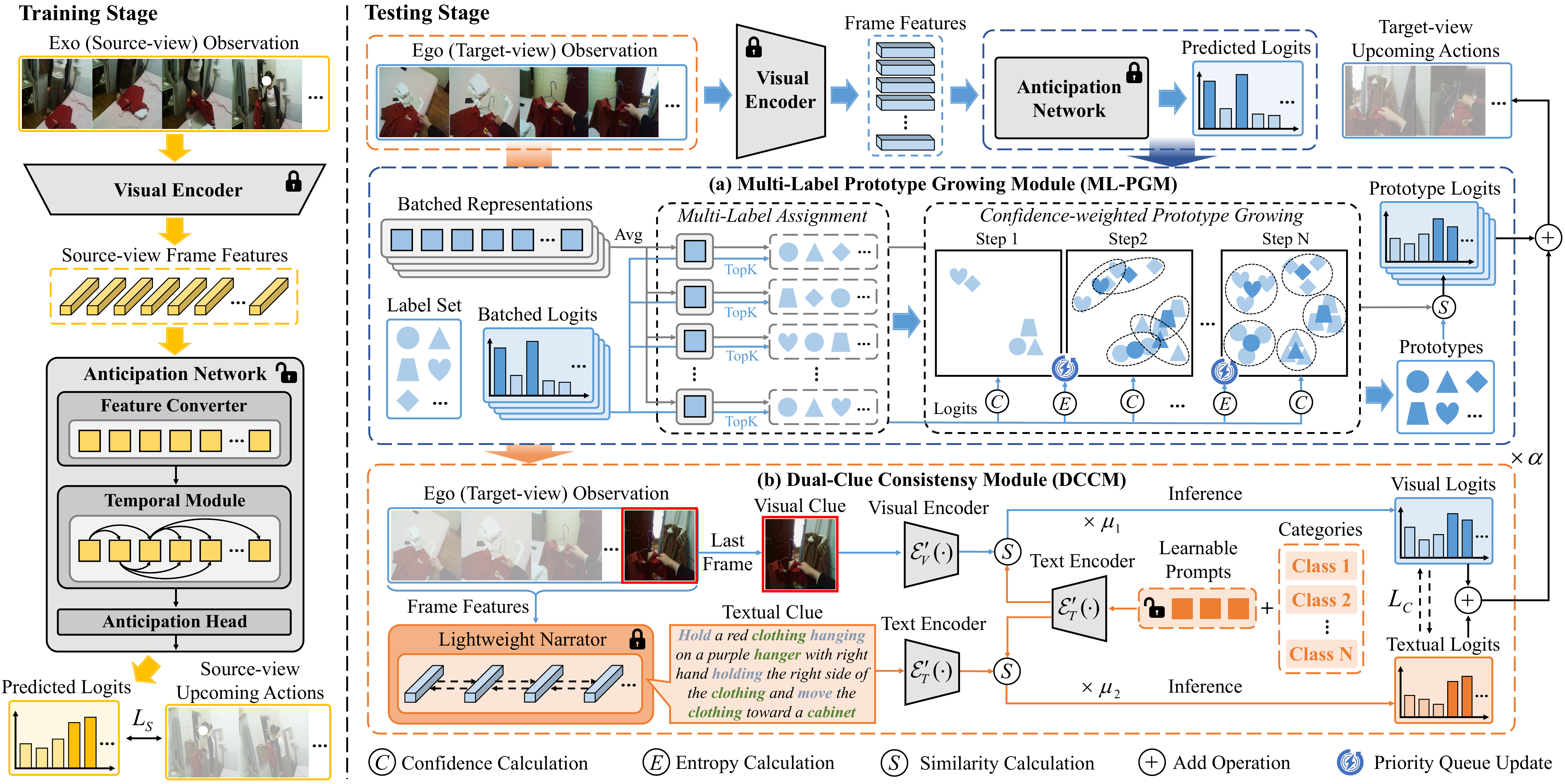}
\caption{Overview of our DCPGN. The model is trained using the source-view data. During testing, the ML-PGM assigns each representation to multiple labels, then performs confidence-based reweighting under an entropy priority queue strategy to balance multiple positive classes. Moreover, to complement the visual clue containing various objects, the DCCM incorporates a lightweight narrator to generate descriptions for action progressions with nouns (green) and verbs (purple). We integrate cross-modality clues by constraining their logits for temporally and spatially bridging the view gap. Finally, the prototype, visual, and textual logits are added for target-view anticipation.}
\label{fig:2}
\end{figure*}

\subsection{Ego-Exo Cross-view Understanding}
Understanding the ongoing or upcoming activities across Ego and Exo views is essential for AI \cite{nagarajan2021shaping,li2025challenges} to emulate human imitation progress, and many Ego-Exo datasets \cite{de2009guide,jia2020lemma,rai2021home,kwon2021h2o,sener2022assembly101,sigurdsson2018charades} have been proposed. EgoExo-4D \cite{grauman2024ego} is currently the largest dataset with multi-view videos. EgoExo-Fitness \cite{li2024egoexo} provides whole-body videos in fitness scenes. EgoExoLearn \cite{huang2024egoexolearn} contains asynchronous videos and introduces the cross-view action anticipation task. EgoMe \cite{qiu2025egome} consists of paired Ego-Exo videos with detailed annotations. Inspired by EgoExoLearn, we construct a brand-new \textit{EgoMe-anti} to support our research.

Meanwhile, many works associate Ego and Exo videos to learn view-unified representations \cite{ardeshir2018exocentric,xue2023learning,li2021ego,luo2025viewpoint} for various downstream tasks, such as cross-view association \cite{xu2024retrieval,huang2025sound}, action recognition \cite{truong2025cross,huang2021holographic,li2021ego,park2025bootstrap}, video captioning \cite{ohkawa2023exo2egodvc,shi2025unsupervised}, and action anticipation or planning \cite{huang2024egoexolearn,zhang2025exo2ego}. The corresponding methods can be summarized in three main directions: cross-view co-training \cite{ohkawa2023exo2egodvc,xu2024retrieval,wang2023learning}, unsupervised adaptation \cite{shi2025unsupervised,xia2022incomplete}, and knowledge distillation \cite{li2021ego,radevski2023multimodal,shi2024cognition}. Quattrocchi et al. \cite{quattrocchi2024synchronization} have proposed an excellent method, which uses labeled Exo data and unlabeled Exo-Ego videos for cross-view transferring with outstanding performance. Unlike the above works requiring target-view data during training, our TE$^{2}$A$^{3}$ task aims at efficient online test-time adaptation with unlabeled target-view data.

\subsection{Test-Time Adaptation}
Test-Time Adaptation (TTA) aims to adapt to the target distribution by adjusting the model online during the testing stage with unlabeled data. Tent \cite{wang2020tent} is a pioneering method that adjusts the batch normalization (BN) parameters by entropy optimization. LAME \cite{boudiaf2022parameter} adopts Laplacian adjusted maximum likelihood estimation to tune the output. Then, lots of works integrate TTA with CLIP \cite{radford2021learning}. Some works \cite{liu2024dart,ma2023swapprompt,dobler2024lost,zhang2024robust,abdul2023align} introduce learnable prompts. For example,  TPT \cite{shu2022test} uses prompts and minimizes the entropies of augmented views, and Diff-TPT \cite{feng2023diverse} improves it by leveraging diffusion models \cite{mei2025efficient,10081412}. Some works consider establishing caches \cite{zhang2022tip,karmanov2024efficient,iwasawa2021test,zhang2024boostadapter} or prototypes \cite{zhang2024dual,liang2025advancing} during test time, which dynamically maintain the historical samples with high confidence. ZERO \cite{farina2024frustratingly} is a simple yet effective method that sets the softmax temperature to zero to convert the probabilities into one-hot results. ML-TTA \cite{wu2025multi} achieves multi-label TTA via bound entropy minimization, whereas it is for image-level classification rather than Ego-Exo video-level anticipation in our TE$^{2}$A$^{3}$ task.

Recently, TTA technology has been extended to fields such as image segmentation \cite{shin2022mm}, image restoration \cite{gou2023test}, object detection \cite{wang2025efficient}, video classification \cite{lin2023video}, depth completion \cite{park2024test}, cross-modal retrieval \cite{li2024test}, and so on. Unlike these works, we adapt between the Ego and Exo videos with a significant inter-view gap for the action anticipation task.

\section{Method}
To address the TE$^{2}$A$^{3}$ task, we propose a Dual-Clue enhanced Prototype Growing Network (DCPGN) as shown in Fig. \ref{fig:2}. In the training stage, the model is trained via a BCE loss denoted as $L_S$ relying on the source-view labeled data. Following the common setting \cite{huang2024egoexolearn}, we utilize a frozen CLIP \cite{radford2021learning} as the visual encoder to extract the frame features, and a trainable TA3N \cite{Chen_2019_ICCV} to anticipate the upcoming actions. During testing, we first propose a Multi-Label Prototype Growing Module (ML-PGM), which progressively accumulates multi-label knowledge to learn unbiased prototypes to balance multiple positive classes. Then, the Dual-Clue Consistency Module (DCCM) constructs consistency between the visual and textual clues for explicitly bridging the Ego-Exo temporal-spatial inter-view gap.

\subsection{Preliminaries}
\subsubsection{Action Anticipation}
Action anticipation aims to forecast the upcoming fine-level actions based on the observation video, with many potential applications such as AI assistance \cite{Lu2024AIAF}. Following the common settings \cite{huang2024egoexolearn,damen2022rescaling}, we represent the actions with noun and verb classes. In detail, given an action event denoted as $e=({{t}_{s}},{{t}_{e}},c)$, where ${{t}_{s}}$, ${{t}_{e}}$, $c$ denote the start timestamp, end timestamp, and classes, respectively. The model can only access an ${{\tau }_{o}}$ second observation video clip $O$ from the time of ${{t}_{s}}-({{\tau }_{i}}+{{\tau }_{o}})$ to ${{t}_{s}}-{{\tau }_{i}}$ to anticipate the classes $c=\{{{y}_{i}}\}_{i=1}^{{{N}_{u}}}\in \mathcal{C}$, where ${\tau}_{i}$ is the interval time, ${N}_{u}$ denotes the number of upcoming actions, and $\mathcal{C}$ means the class set.

\subsubsection{Problem Definition for the TE$^{2}$A$^{3}$ task}
We make the first exploration of the TE$^{2}$A$^{3}$ task, which adjusts the source-view-trained model during test time for action anticipation in the target view. We define one of the annotated Ego or Exo views as the ``source view", while the other unlabeled view as the ``target view". We first perform supervised training based on the source-view dataset ${\mathcal{D}_{S}}=\{(O_{i}^{S},Y_{i}^{S})\}_{i=1}^{{{N}_{s}}}$, where $O_{i}^{S}$, $Y_{i}^{S}$ denote the source-view observations and ground-truth labels, and ${N}_{s}$ means the number of data. The trained model can be denoted as $\mathcal{M}_{S}$. During the test time, the model $\mathcal{M}_{S}$ is real-time adjusted by a online stream of target-view data ${{\mathcal{D}}_{T}}=\{O_{j}^{T}\}_{j=1}^{{{N}_{t}}}$ to obtain ${\mathcal{M}_{S\to T}}$, which encourages the model to adapt to target-view videos to accurately anticipate the upcoming actions.

\subsection{Multi-Label Prototype Growing Module}

\begin{algorithm}[!t]
\renewcommand{\algorithmicrequire}{\textbf{Input:}}
\renewcommand{\algorithmicensure}{\textbf{Output:}}
\caption{Multi-Label Prototype Growing Module}
\label{alg:1}
\begin{algorithmic}[1]
\REQUIRE A source-view model ${{\mathcal{M}}_{S}}=\{{{\mathcal{E}}},{{\mathcal{A}}_{S}}\}$, observation video batches $\{{{\mathcal{O}}_{1}},\cdots {{\mathcal{O}}_{N_b}}\}$, where $\mathcal{O}_{n}=\{O_{j}^{T}\}_{j=1}^{bs}$, number of classes $C'$, and hyperparameters $K$ and $N$;
\ENSURE Target-view prototype classifier $P^T$;
\STATE Initialize $C'$ class memory banks $\{{{\mathcal{B}}_{i}}\}_{i=1}^{{{C}'}}$ as empty;
\FOR{batch=1 \textbf{to} $N_b$}
\FOR{$j$=1 \textbf{to} $bs$} 
\STATE $\{\bar{f}_{v,j}^{T};L_{j}^{T}\}\leftarrow {{\mathcal{A}}_{S}}(\mathcal{E}(O_{j}^{T}))$;  //  \text{ }Extract video-level representation $\bar{f}_{v,j}^{T}$ and predict initial logits $L_{j}^{T}$
\STATE $\{Y_{K,j}^{T};L_{K,j}^{T}\}\leftarrow L_{j}^{T}$; \text{ } // \text{ }Assign \textit{Top-K} pseudo labels ${{Y}_{K,j}^{T}}$ and obtain the confidence scores $L_{K,j}^{T}$ of the $j$-th sample in the batch as in Eq. (\ref{eq:1})
\STATE $\mathcal{H}_{j}^{T}\leftarrow En(L_{j}^{T})$; // Calculate entropy as in Eq. (\ref{eq:3})
\ENDFOR
\STATE $\{\mathcal{B}_{i}^{new}\}_{i=1}^{{{C}'}}\leftarrow \{{{\mathcal{B}}_{i}}\}_{i=1}^{{{C}'}}$;\text{ }\text{ }\text{ } // \text{ }Update memory banks (maximum capacity $N$) via entropy as in Eq. (\ref{eq:4})
\STATE ${{P}^{T}}\leftarrow \{\mathcal{B}_{i}^{new}\}_{i=1}^{{{C}'}}$; // Perform confidence-based reweighting to obtain prototype classifier in Eq. (\ref{eq:5})
\ENDFOR
\end{algorithmic}
\end{algorithm}

Most of the current TTA works adopt entropy-based optimization \cite{wang2020tent,farina2024frustratingly} or cache construction \cite{karmanov2024efficient,shu2022test} methods. Although effective in single-label tasks, they cause biased predictions towards the most confident class and yield unsatisfactory results in our TE$^{2}$A$^{3}$ task, which involves multiple action candidates. Therefore, as shown in panel (a) of Fig. \ref{fig:2}, we propose a Multi-Label Prototype Growing Module (ML-PGM). It first assigns multiple pseudo labels to the extracted representations and then performs confidence-based reweighting. This process is updated via an entropy priority queue strategy to balance multiple positive classes. The overall progress of ML-PGM is shown in Algorithm \ref{alg:1}.

The source-view model is denoted as ${{\mathcal{M}}_{S}}=\{{{\mathcal{E}}},{{\mathcal{A}}_{S}}\}$, where $\mathcal{E}$ is the fixed visual encoder and ${\mathcal{A}}_{S}$ is the anticipation network. Given a target-view observation ${{O}^{T}}$, we first utilize $\mathcal{E}$ to extract the frame features ${{F}^{T}}\in {{\mathbb{R}}^{L\times C}}$, where $L$ denotes the number of sampled frames and $C$ is the feature dimension. Then, we feed the ${F}^{T}$ into ${\mathcal{A}}_{S}$ to obtain the intermediate representations ${{\bar{F}}^{T}}$ and the initial predicted logits ${{L}^{T}}=\{l_{i}^{T}\}_{i=1}^{{{C}'}}$, where ${C}'$ is the total number of classes. Next, we average ${{\bar{F}}^{T}}$ to obtain the video-level representation $\bar{f}_{v}^{T}\in {{\mathbb{R}}^{C_1}}$. Moreover, we assign the \textit{Top-K} classes as pseudo labels to the video-level representation and obtain the corresponding confidence scores based on the logits:
\begin{equation}
\{Y_{K}^{T};L_{K}^{T}\}=\{{{y}_{c}};l_{c}^{T}|c\in \mathcal{C},\mathbb{I}(Ran{{k}_{\downarrow }}({{L}^{T}})\le K)\}
\label{eq:1}
\end{equation}
where ${{Y}_{K}^{T}}$ denotes the assigned \textit{Top-K} pseudo labels, and $L_{K}^{T}$ is the corresponding confidence scores. $\mathbb{I}(\cdot )$ denotes the indicator function, and $Ran{{k}_{\downarrow }}(\cdot )$ represents the decreasing ranking. To measure the reliability of the pseudo labels, we also compute the entropy of predicted logits as follows:
\begin{equation}
{{\mathcal{H}}^{T}}=En(L^T)=-\sum\limits_{i=1}^{{{C}'}}{\sigma (l_{i}^{T})}\log \sigma (l_{i}^{T})
\label{eq:3}
\end{equation}
where ${{\mathcal{H}}^{T}}$ is the entropy and $\sigma (\cdot )$ means softmax function. Hence, we obtain a data tuple $\mathcal{T}=(\bar{f}_{v}^{T},{{Y}_{K}^{T}},L_{K}^{T},{{\mathcal{H}}^{T}})$.

We initialize the memory banks $\{{{\mathcal{B}}_{i}}\}_{i=1}^{{{C}'}}$ for each class. When a batch of data tuples $\{{{\mathcal{T}}_{j}}\}_{j=1}^{bs}$ arrives, an entropy priority queue strategy is applied. Specifically, for the $j$-th data in the current batch, consider a representation $\bar{f}_{v,j}^{T}$ with the class-$i$ pseudo label ${{y}_{i,j}}\in {{Y}_{K,j}^{T}}$, if ${{\mathcal{B}}_{i}}$ is not full, we directly add the $\bar{f}_{v,j}^{T}$, confidence $l_{i,j}^{T}\in L_{K,j}^{T}$, and entropy $\mathcal{H}_{j}^{T}$ to ${{\mathcal{B}}_{i}}$. If ${{\mathcal{B}}_{i}}$ is full, we retain the corresponding representations and confidence scores with the lowest $N$ entropies, where $N$ is the maximum capacity. This process can be formulated as:
\begin{equation}
\mathcal{B}_{i}^{new} = 
\begin{cases} 
{{\mathcal{B}}_{i}}\cup \{\bar{f}_{v,j}^{T},l_{i,j}^{T},\mathcal{H}_{j}^{T}\} \text{ }\text{ }\text{ }\text{ }\text{ }\text{ }\text{ }\text{ }\text{ }\text{ }\text{ }\text{ }\text{ }\text{ }\text{ }\text{ }\text{ }\text{ }\text{ }\text{ }\text{ }\text{ }\text{ }\text{ }\text{if}\text{ }  |{{\mathcal{B}}_{i}}|<N \\
\{\bar{f}_{v}^{T},l_{i}^{T},{{\mathcal{H}}^{T}}\}\in \{{{\mathcal{B}}_{i}}\cup \{\bar{f}_{v,j}^{T},l_{i,j}^{T},\mathcal{H}_{j}^{T}\} \\ \text{ }\text{ }\text{ }\text{ }\text{ }\text{ }\text{ }\text{ }\text{ }\text{ }\text{ }|\text{ }\mathbb{I}(Ran{{k}_{\uparrow }}({{\mathcal{H}}^{T}})\le N)\} \text{ }\text{ }\text{ }\text{ }\text{ }\text{ }\text{ }\text{ }\text{if}\text{ }|{{\mathcal{B}}_{i}}|=N
\end{cases}
\label{eq:4}
\end{equation}
Since lower entropy implies lower prediction uncertainty \cite{zhang2024dual}, the representations and pseudo labels in memory banks become more reliable over time during the test stage.

However, in some circumstances, a few negative classes with low confidence may be included in the \textit{TopK} classes and thus incorrectly assigned as pseudo labels during updating. To mitigate such interference, we leverage the confidence to reweight the representations in the memory bank to calculate the prototypes. Assume that the updated memory bank of class $i$ is denoted as ${{\mathcal{B}}_{i}^{new}}=\{\bar{f}_{v,k}^{T},l_{i,k}^{T},\mathcal{H}_{k}^{T}\}_{k=1}^{{N}'\le N}$, where $k$ denotes the data index in $\mathcal{B}_{i}^{new}$, and $N'$ is the number of stored representations. For class $i$, we weight the video-level representations $\bar{f}_{v,k}^{T}$ to calculate its prototype:
\begin{equation}
p_{i}^{T}=\sum\limits_{k=1}^{N'}{\eta (l_{i,k}^{T})\cdot \bar{f}_{v,k}^{T}}
\label{eq:5}
\end{equation}
$\eta (\cdot )$ is a normalization operation. The prototype classifier is denoted as ${{P}^{T}}\in {{\mathbb{R}}^{{C}'\times C_1}}$, and we compute the similarity between ${{P}^{T}}$ and $\bar{f}_{v}^{T}$ to get the prototype logits ${{L}_{p}} \in {{\mathbb{R}}^{{C}'}}$.

\subsection{Dual-Clue Consistency Module}
Due to inconsistent spatial objects and asynchronous temporal progressions, the Ego and Exo videos exhibit a significant inter-view gap, which prevents the source-view model from maintaining its learned mapping from observation to the target-view upcoming actions. Moreover, the ML-PGM suffers from the inevitable prototype instability in the initial few adaptation steps. Therefore, as shown in panel (b) of Fig. \ref{fig:2}, we design a simple yet effective Dual-Clue Consistency Module (DCCM). Specifically, we first introduce a lightweight narrator to generate textual clues to complement visual clues. Then, their inferred logits are constrained to construct the dual-clue consistency for explicitly bridging the Ego-Exo gap in temporal and spatial dimensions.

The observation videos contain abundant clues, such as potential interactive objects and ongoing activities, which are highly relevant to the upcoming actions. Therefore, we first extract the final frame of the given observation as the visual clue ${{\bar{V}}_{C}}$, which contains various visual objects in the scene. Nevertheless, although the visual clue provides rich spatial information, it is insufficient to establish correlations between clues and upcoming actions because a single frame cannot reflect the temporal progression. Therefore, we introduce an extra lightweight narrator, which models the dependencies between observation frames and generates view-agnostic descriptions, naturally serving as indicators for temporal activities in progress. Thanks to open-source datasets \cite{grauman2024ego,qiu2025egome,huang2024egoexolearn} with fine-level language annotations, we first collect a series of video-text pairs $\{{{V}_{i}},{{T}_{i}}\}_{i=1}^{N}$ from existing datasets. Then, we use the frozen visual encoder $\mathcal{E}$ to extract the frame features $F_{V}$ for each video. Next, we leverage the frame features and their captions to train a universal narrator $\mathcal{N}(\cdot )$ based on GRU units with an attention mechanism for computational efficiency, whose details are available in the Supplementary Material. During testing, we use the frozen narrator to generate textual clues, as follows:
\begin{equation}
{{\bar{T}}_{C}}=\{{{w}_{1}},{{w}_{2}},\cdots {{w}_{{{L}_{t}}}}\}=\mathcal{N}({{F}^{T}})
\end{equation}
where ${{\bar{T}}_{C}}$ denotes the description as text clue, $w_i$ means the generated words, and $L_t$ is the length of the textual clue.

Inspired by previous works \cite{shi2024cognition, zhong2022regionclip}, we adopt the on-the-shelf CLIP \cite{radford2021learning}, which is pre-trained on the world-scale paired data to correlate the mined clues and action class labels. We denote the CLIP visual encoder as ${\mathcal{E}'_{V}}(\cdot )$ and the CLIP textual encoder is ${\mathcal{E}'_{T}}(\cdot )$. The features of the visual clue ${{\bar{V}}_{C}}$ and the textual clue ${{\bar{T}}_{C}}$ are extracted as follows:
\begin{equation}
{{\bar{f}}_{v}}=\mathcal{E}'_{V}({{\bar{V}}_{C}})\text{    }\text{    }\text{    }\text{    }\text{    }\text{    }\text{    }\text{    }\text{    }\text{    }{{\bar{f}}_{t}}=\mathcal{E}'_{T}({{\bar{T}}_{C}})
\end{equation}
where ${{\bar{f}}_{v}}\in {{R}^{C}}$ and ${{\bar{f}}_{t}}\in {{R}^{C}}$ denote the extracted visual and textual clue features, respectively. In addition, we insert the learnable prompts ${{p}_{l}}\in {{R}^{{{L}_{p}}\times C}}$ into the embeddings of the action classes and extract the corresponding features:
\begin{equation}
{{\bar{T}}_{cls}}=\{[{{p}_{l}};\text{Embed}({{y}_{c}})]\text{ }|\text{ }c\in \mathcal{C}\}
\end{equation}
\begin{equation}
{{\bar{F}}_{l}}=\{{{\bar{f}}_{l,1}},\cdots {{\bar{f}}_{l,{C}'}}\}={\mathcal{E}'_{T}}({{\bar{T}}_{cls}})
\end{equation}
where $[;]$ means the concatenation operation, $\text{Embed}(\cdot )$ is the CLIP embedding operation, and ${{\bar{F}}_{l}}=\{{{\bar{f}}_{l,i}}\}_{i=1}^{{{C}'}}\in {{\mathbb{R}}^{{C}'\times C}}$ denotes the learnable features of action classes. 

Then, we calculate the similarity between the visual or textual clue features and the learnable class features to infer their respective logits, which can be formulated as follows:
\begin{equation}
{{l}_{v,i}}=\frac{{{{\bar{f}}}_{v}}\cdot {{({{{\bar{f}}}_{l,i}})}^{T}}}{||{{{\bar{f}}}_{v}}|{{|}_{2}}\cdot ||{{{\bar{f}}}_{l,i}}|{{|}_{2}}}\cdot {{\mu }_{1}}\text{    }\text{    }{{l}_{t,i}}=\frac{{{{\bar{f}}}_{t}}\cdot {{({{{\bar{f}}}_{l,i}})}^{T}}}{||{{{\bar{f}}}_{t}}|{{|}_{2}}\cdot ||{{{\bar{f}}}_{l,i}}|{{|}_{2}}}\cdot {{\mu }_{2}}
\end{equation}
where ${{L}_{v}}=\{{{l}_{v,i}}\}_{i=1}^{{{C}'}}$ and ${{L}_{t}}=\{{{l}_{t,i}}\}_{i=1}^{{{C}'}}$ denote the inferred visual and textual logits, and ${{\mu }_{1}}$, ${{\mu }_{2}}$ are scaling hyperparameters. Next, we use the softmax normalization to obtain the probability distribution responses $P_v$ and $P_t$ for $L_v$ and $L_t$, respectively. And we constrain the inferred logits from cross-modality clues through a Kullback-Leibler (KL) divergence to construct the dual-clue consistency for a comprehensive temporal-spatial correlation between the mined clues and action classes to be anticipated. The consistency loss $L_C$ can be computed as follows:
\begin{multline}
{{L}_{C}}=KL({{P}_{v}}||{{P}_{t}})+KL({{P}_{t}}||{{P}_{v}})=\\
\sum\limits_{i}^{{{C}'}}{{{P}_{v}}(i)\log \frac{{{P}_{v}}(i)}{{{P}_{t}}(i)}}+\sum\limits_{i}^{{{C}'}}{{P}_{t}}(i)\log \frac{{{P}_{t}}(i)}{{{P}_{v}}(i)}
\end{multline}

Finally, we weighted add the visual logits $L_v$, and textual logits $L_t$, and the aforementioned prototype logits $L_p$ as the target-view anticipation result ${{L}_{final}}$, as follows:
\begin{equation}
{{L}_{final}}={{L}_{p}}+\alpha \cdot ({{L}_{v}}+{{L}_{t}})
\end{equation}
where $\alpha$ is a hyperparameter to balance the distinct logits.

\section{Experiments}

\begin{table*}[!t]
\centering
\caption{Quantitative results on the \textit{EgoMe-anti} and \textit{EgoExoLearn} benchmarks under the Exo2Ego and Ego2Exo settings.}
\label{tab:1}
\scalebox{0.92}{
\begin{tabular}{p{3.7cm}<{\raggedright}|p{1.2cm}<{\centering}p{1.2cm}<{\centering}|p{1.2cm}<{\centering}p{1.2cm}<{\centering}|p{1.2cm}<{\centering}p{1.2cm}<{\centering}|p{1.2cm}<{\centering}p{1.2cm}<{\centering}}
\toprule
\multirow{3}{*}{Methods} & \multicolumn{4}{c|}{\textit{EgoMe-anti}}  & \multicolumn{4}{c}{\textit{EgoExoLearn}}                                                         \\ 
     & \multicolumn{2}{c}{Exo2Ego} & \multicolumn{2}{c|}{Ego2Exo}   & \multicolumn{2}{c}{Exo2Ego}     & \multicolumn{2}{c}{Ego2Exo}     \\ \cmidrule{2-9} 
     & Noun & Verb & Noun & Verb & Noun & Verb & Noun & Verb \\ \midrule
      \rowcolor{gray!15} Ours without Adaptation & 71.94 & 32.46 & 64.24  & 30.07 & 31.91	& 34.36 & 35.28& 33.03 \\ 
     Tent \cite{wang2020tent}  &74.14	&35.63& 68.50&	32.32& 34.71&	36.22& 38.61	&35.04 \\
     TPT \cite{shu2022test} & 74.88	& 35.97 & 68.65 & 	32.55 & 35.29	& 36.48 &38.82 &	35.14 \\
     VITTA \cite{lin2023video}  &  75.65 & 	35.94 &  68.68	&  33.84 &  35.06 & 	36.44 & 39.17& 	35.65 \\
     TDA \cite{karmanov2024efficient} &76.96&	36.66& \underline{69.52}&	33.45& 36.11&	36.79 &39.81	&35.11 \\
     ZERO \cite{farina2024frustratingly}  & 75.19 &	36.03 &69.12&	33.30 &\underline{37.19}&	36.76 &40.27&	36.73\\
     TCA \cite{wang2024less}  & \underline{77.23}&	36.33& 69.27&	34.08& 35.00&	36.95& 38.84&	36.18 \\
     ML-TTA \cite{wu2025multi}  & 77.11	&\underline{36.92} & 69.46	&\underline{34.39} & 36.35	&\underline{37.67} & \underline{42.96}&	\underline{40.43} \\
     \textbf{DCPGN (Ours)} & \textbf{79.03}&	\textbf{43.84 }&\textbf{72.01}	&\textbf{40.10} &\textbf{46.26}	&\textbf{42.98} & \textbf{48.48}&	\textbf{46.51} \\            
\bottomrule
\end{tabular}}
\end{table*}

\subsection{Experimental Settings}
\subsubsection{Benchmarks and Evaluation Metrics}
We construct a novel \textbf{\textit{EgoMe-anti}} benchmark based on the EgoMe \cite{qiu2025egome} dataset. Specifically, EgoMe contains 7902 video pairs of Exo observer and Ego follower with a total duration of 82 hours, including detailed annotations such as fine-level timestamps and descriptions. However, the original annotations are raw descriptive sentences rather than fine-grained noun or verb classes. Therefore, we first filter the videos and only retain the correctly following Ego-Exo pairs. Then, we use the Spacy library to extract verbs and nouns from sentences, transform them into base form, and perform synonym fusion and replacement. Finally, we remove low-frequency (below 0.5$\%$) and anomalous classes. Meanwhile, Ego and Exo data are constrained to share identical label sets. We also evaluate our method on \textbf{\textit{EgoExoLearn}} \cite{huang2024egoexolearn}. EgoExoLearn is for bridging asynchronous Ego and Exo procedural activities. It contains 120 hours of videos on daily and professional scenes. In addition, it provides annotations such as fine and coarse-level noun/verb classes, timestamps, and descriptions. We perform experiments on its official action anticipation benchmark. 

Following common settings \cite{huang2024egoexolearn,damen2022rescaling}, we define two settings: \textbf{Exo2Ego} and \textbf{Ego2Exo} to predict the noun/verb classes, and use the class-mean Top-5 recall for evaluation.

\subsubsection{Implementation Details}
Following the pioneering work \cite{huang2024egoexolearn}, we use CLIP (ViT-L/14) \cite{radford2021learning} as the visual encoder. The observation time ${\tau}_o$ and interval time ${\tau}_i$ are set to 2s and 1s, respectively. For an observation video, we uniformly sample 5 frames as input. The feature dimension $C$ of the visual encoder is 768, and $C_1$ of the anticipation network is set to 512. The hyperparameter $K$ is set to 5 for \textit{EgoExoLearn} benchmark, while $K$ is 3 for \textit{EgoMe-anti}, and the maximum capacity $N$ of a memory bank is set to 500. The scaling parameters ${{\mu }_{1}}$, ${{\mu }_{2}}$ are set to 1.0 and 0.5 respectively and $\alpha$ is 0.5. The length of the learnable prompts $L_p$ is set to 4. Moreover, the narrator $\mathcal{N}(\cdot )$ is implemented following the official video-caption.pytorch library. In the testing stage, the batch size is set to 64 without any data augmentation strategies. The learnable prompts are optimized online by the SGD \cite{bottou2010large} optimizer, and the learning rate is 1e-4 for \textit{EgoExoLearn} and 5e-4 for \textit{EgoMe-anti} due to the difference in dataset scale.

\begin{table*}[!t]
\centering
\caption{Ablation results under the Exo2Ego and Ego2Exo benchmarks on the \textit{EgoMe-anti} and \textit{EgoExoLearn} benchmarks. $L_C$ is the dual-clue consistency loss. ``VC" and ``TC" denote the visual clues and textual clues, respectively. ``Conf" represents the confidence-based reweighting, and ``ML" denotes the multi-label assignment. ``S" means only assign a single class as the pseudo label for the representation.}
\label{tab:2}
\scalebox{0.93}{
\begin{tabular}{p{0.9cm}<{\centering}p{0.9cm}<{\centering}p{0.9cm}<{\centering}p{0.9cm}<{\centering}p{0.9cm}<{\centering}|p{0.9cm}<{\centering}p{0.9cm}<{\centering}|p{0.9cm}<{\centering}p{0.9cm}<{\centering}|p{0.9cm}<{\centering}p{0.9cm}<{\centering}|p{0.9cm}<{\centering}p{0.9cm}<{\centering}}
\toprule
\multirow{3}{*}{$L_C$} &\multirow{3}{*}{VC} &\multirow{3}{*}{TC} &\multirow{3}{*}{Conf} &\multirow{3}{*}{ML} &  \multicolumn{4}{c|}{\textit{EgoMe-anti}}  & \multicolumn{4}{c}{\textit{EgoExoLearn}}                                                         \\ 
     & & & &  &\multicolumn{2}{c}{Exo2Ego} & \multicolumn{2}{c|}{Ego2Exo}   & \multicolumn{2}{c}{Exo2Ego}     & \multicolumn{2}{c}{Ego2Exo}     \\ \cmidrule{6-13} 
     & & & &  & Noun & Verb & Noun & Verb & Noun & Verb & Noun & Verb \\ \midrule
     \checkmark & \checkmark & \checkmark & \checkmark &\checkmark & \textbf{79.03}&	\textbf{43.84 }&\textbf{72.01}	&\textbf{40.10} &\textbf{46.26}	&\textbf{42.98} & \textbf{48.48} &	\textbf{46.51} \\ 
      & \checkmark & \checkmark & \checkmark &\checkmark & 78.67	& 42.72 & 71.29	 &39.20 & 44.80	& 42.73 & 47.97 &	45.84 \\ 
       & & \checkmark & \checkmark &\checkmark & 76.92 &	42.37 & 69.65 &	37.99 & 41.32 &	42.25 &	45.73 &	45.23 \\ 
        & \checkmark & & \checkmark &\checkmark &  77.56 &	42.15 & 70.47 &	37.66 & 41.94	& 41.93 &	 45.88 & 43.72 \\ 
     \cmidrule{1-13}
         & & & \checkmark &\checkmark & 76.11	& 40.84 & 69.21 & 37.55 & 38.43 &	40.54  &	44.53 &		43.25 \\
        & & & &\checkmark & 74.63 &	39.87 & 68.76 &	36.84 &37.76&	39.33 &43.47	&42.01 \\
        & & & & S & 72.74 & 35.47 & 65.51 &	32.34 &
        34.70 &	34.98 & 38.49 & 37.71 \\
        & & & &  &  71.94 & 32.46 & 64.24 & 30.07  & 31.91  &	34.36  &35.28  &33.03 \\
\bottomrule
\end{tabular}}
\end{table*}

\subsection{Comparison with State-of-the-Art Methods}
We compare our DCPGN with many image-level or video-level TTA methods, i.e., Tent \cite{wang2020tent}, TPT \cite{shu2022test}, VITTA \cite{lin2023video}, TDA \cite{karmanov2024efficient}, ZERO \cite{farina2024frustratingly}, TCA \cite{wang2024less}, and ML-TTA \cite{wu2025multi}. We re-implement these methods according to their officially released codes. Moreover, we conduct experiments under the same settings and backbone to ensure a fair comparison.

The left panel in Table \ref{tab:1} shows the performance on the \textit{EgoMe-anti} benchmark, and the results show that our DCPGN significantly outperforms the other related TTA methods. For the Exo2Ego setting, DCPGN surpasses the second-place TCA \cite{wang2024less} by 1.80$\%$ in the noun anticipation, and outperforms the recent ML-TTA \cite{wu2025multi} by 6.92$\%$ in verb. For the Ego2Exo setting, which also has many applications such as multi-robot collaboration, our DCPGN also yields performance gains of 2.49$\%$ and 5.71$\%$ in the noun and verb anticipation, respectively. We also evaluate our method on the common \textit{EgoExoLearn} benchmark. The results in the right panel in Table \ref{tab:1} demonstrate the superiority of our method compared with other TTA methods. The recently proposed ML-TTA \cite{wu2025multi} utilizes the label binding strategy for the multi-label image-level TTA and achieves a high performance on \textit{EgoExoLearn}. However, our DCPGN explicitly bridges the significant inter-view gap in temporal and spatial dimensions for video-level action anticipation. It remarkably outperforms ML-TTA by 9.91$\%$ and 5.31$\%$ in the noun/verb anticipation under the Exo2Ego setting, and by 5.52$\%$ and 6.08$\%$ under the Ego2Exo setting. The above results demonstrate the effectiveness of our method in balancing multiple positive classes and bridging the Ego and Exo videos in temporal and spatial dimensions.

\subsection{Ablation Study}
In Table \ref{tab:2}, we evaluate the key components of our DCPGN. In the second row, we remove the dual-clue consistency loss $L_C$, exhibiting a 0.36$\%$, 1.12$\%$, 0.72$\%$, 0.90$\%$ drop on \textit{EgoMe-anti}, and 1.46$\%$, 0.25$\%$, 0.51$\%$, 0.67$\%$ on \textit{EgoExoLearn}, showing its effectiveness in correlating the cross-modality clues. In the third and fourth rows, we remove the visual and textual clues, respectively, and all metrics decrease compared to the second row. Furthermore, the results of noun anticipation decrease more when removing the visual clues, while removing the textual clues leads to a more significant drop in verb anticipation. Such results indicate that visual clues mainly contain spatial object information, while textual clues indicate temporal dependencies of activities. Next, we remove the entire DCCM, and the results under all settings drop remarkably, demonstrating its effectiveness in bridging the significant Ego-Exo inter-view gap.

Then, we evaluate the ML-PGM in rows 5-8. In the sixth row, we remove the confidence-based reweighting operation in the ML-PGM, leading to a decrease of 1.48$\%$, 0.97$\%$, 0.45$\%$, 0.71$\%$ on \textit{EgoMe-anti}, and 0.67$\%$, 1.21$\%$, 1.06$\%$, 1.24$\%$ on \textit{EgoExoLearn}, which shows its effectiveness in mitigating the interference of potential negative classes. In addition, we only assign the \textit{Top-1} class rather than \textit{Top-K} classes as the pseudo label in row seven, causing a significant drop of 1.89$\%$, 4.40$\%$, 3.25$\%$, 4.50$\%$, and 3.06$\%$, 4.36$\%$, 4.98$\%$, 4.30$\%$ on the two benchmarks, respectively. It shows the effectiveness of our method to perform multi-label assignment to balance multiple positive classes.

\subsection{In-depth Analysis}

\subsubsection{Analysis of Multi-Label Assignment}

\begin{table}[t]
\centering
\caption{Analysis of the number of assigned pseudo labels $K$.}
\label{tab:3}
\scalebox{0.89}{
\begin{tabular}{p{1.2cm}<{\raggedright}|p{1.2cm}<{\centering}p{1.2cm}<{\centering}|p{1.2cm}<{\centering}p{1.2cm}<{\centering}}
\toprule
\multicolumn{5}{c}{\textit{EgoMe-anti}}    \\ \midrule
\multirow{2}{*}{$K$} & \multicolumn{2}{c}{Exo2Ego}     & \multicolumn{2}{|c}{Ego2Exo}     \\ \cmidrule{2-5} 
                  & Noun & Verb & Noun & Verb \\ \midrule
$K$=1             & 76.12 &39.89  &  69.55	&36.49         \\ 
$K$=2             &77.89	& 42.65 & 70.67	&38.97  \\ 
$K$=3              &  \textbf{79.03}&	\textbf{43.84}  &  \textbf{72.01} &	\underline{40.10}    \\ 
$K$=4              &  \underline{78.95}	& \underline{43.65}  & \underline{71.43}	& \textbf{40.42}   \\ 
$K$=5           &  78.58	&42.98  &71.22&	40.04    \\ 
\bottomrule
\toprule
\multicolumn{5}{c}{\textit{EgoExoLearn}}    \\ \midrule
\multirow{2}{*}{$K$} & \multicolumn{2}{c}{Exo2Ego}     & \multicolumn{2}{|c}{Ego2Exo}     \\ \cmidrule{2-5} 
                  & Noun & Verb & Noun & Verb \\ \midrule
$K$=1          &  41.77&	35.26 & 45.38&	40.08     \\ 
$K$=2              & 45.02 &36.95& 47.29&	40.05     \\ 
$K$=3              & 45.69 &	37.73& \underline{48.15}&	40.23    \\ 
$K$=4              & \underline{46.15} &	\underline{38.07} & 47.98&	\underline{44.58}    \\ 
$K$=5              &  \textbf{46.26} &	\textbf{42.98} &  \textbf{48.48}	& \textbf{46.51}  \\ 
\bottomrule
\end{tabular}}
\end{table}

$K$ is an important hyperparameter, which facilitates the ML-PGM to balance multiple positive classes. Thus, we conduct experiments to analyze the effect of different settings of $K$ in Table \ref{tab:3} on distinct benchmarks. The results show that the best performance is achieved on \textit{EgoMe-anti} when $K$ is 3, while the optimal $K$ is 5 on \textit{EgoExoLearn}. The reason for the different $K$ setting is that the actions in the videos of \textit{EgoExoLearn} tend to be more compact than those in \textit{EgoMe-anti} (i.e., average number of actions per time segment under the same setting: 2.46 vs. 1.70). Hence, we dynamically set the hyperparameter $K$ for different benchmarks. In addition, when the value of $K$ is 1 (i.e., only assign a single label), the performance deteriorates remarkably under all settings. It demonstrates that traditional single-label methods are likely to cause over-confident predictions and introduce noise. We address this problem via multi-label assignment and confidence-based reweighting updated by an entropy priority queue strategy.

\subsubsection{Analysis of Model Complexity}
In Table \ref{tab:4}, we only incorporate the correlation between visual information and learnable textual classes into the source-view model as the baseline, as with many current CLIP-based TTA methods (e.g., TPT \cite{shu2022test}). The baseline model has a total parameter count of 251.18 M and an inference computational complexity of 367.55 GFLOPs. In the second column, we add the Multi-Label Prototype Growing Module (ML-PGM), which only introduces 8.54 M parameters of the memory banks for all classes with negligible computational costs. In addition, because the lightweight narrator takes the extracted frame features as input, it requires low computational complexity and parameters (i.e., 0.03 GFLOPs and 2.38 M), as shown in the third column. And the feature extraction of the generated textual clues introduces 4.06 GFLOPs of computational complexity and 54.04 M of parameters. Considering the performance improvement, the increased computational complexity and parameters of the above key components are tolerable.

\begin{table}[!t]
\centering
\small
\caption{The inference FLOPs and the number of parameters of the baseline model and key components in the proposed method. ``NG" denotes the narration generation, and ``TCFE" represents the textual clue feature extraction.}
\label{tab:4}
\scalebox{0.9}{
\begin{tabular}{p{2.0cm}<{\raggedright}|p{1.2cm}<{\centering}|p{1.3cm}<{\centering}|p{1.2cm}<{\centering}|p{1.2cm}<{\centering}}
\toprule
      & Baseline & ML-PGM & NG & TCFE \\ \midrule
FLOPs (G)  &367.55 &	0.00 	&0.03	&4.06  \\ \hline
$\#$Params (M)& 251.18 &	8.54 &2.38 &	54.04  \\ 
\bottomrule
\end{tabular}}
\end{table}

\begin{figure}[!t]
\centering
\includegraphics[width=1.0\linewidth]{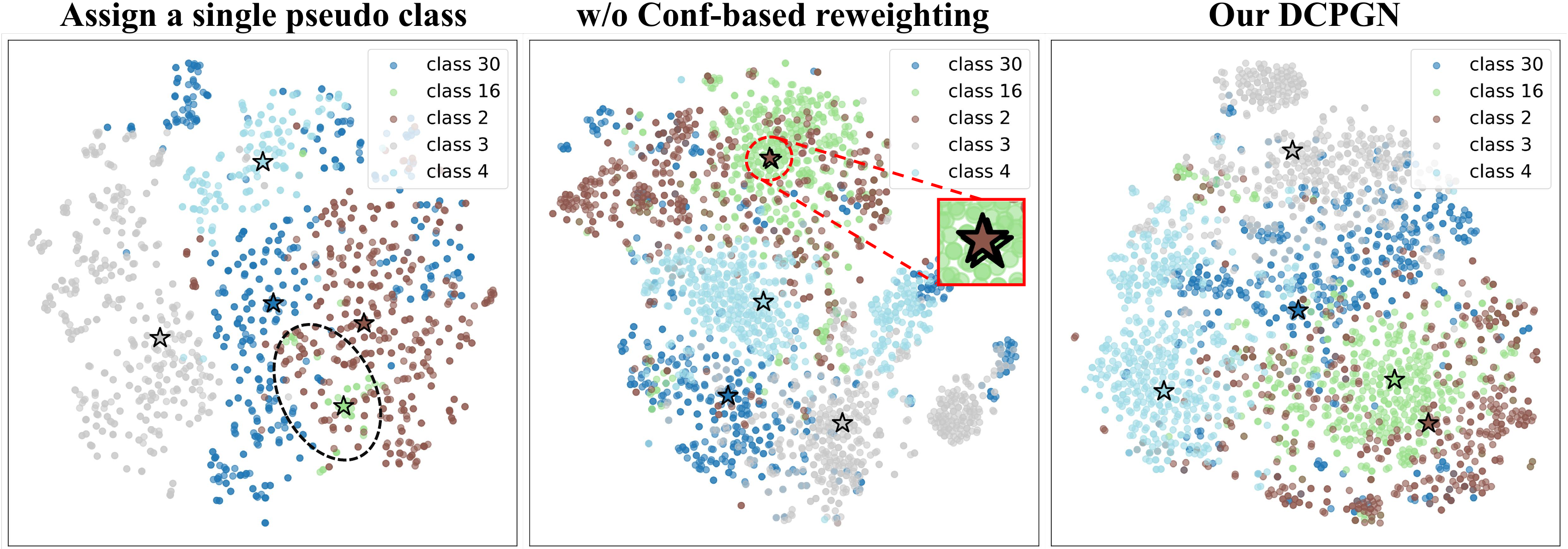}
\caption{Visualization of representations in memory banks and the corresponding prototypes of several classes. Dots denote saved representations and stars denote prototypes (best viewed in color).}
\label{fig:3}
\end{figure}

\subsubsection{Analysis of Representations and Prototypes}
To intuitively visualize the representations in memory banks and the corresponding class prototypes, we use t-SNE \cite{van2008visualizing} to project them into 2-D planes as shown in Fig. \ref{fig:3}. In practice, we select five dominant classes with the most samples and visualize the corresponding representations and prototypes under the setting of the \textit{EgoExoLearn} benchmark.

In the first column, we visualize the representations of the model that only assigns the most confident class. The result shows that there are only a few representations assigned to class 16 (circled in black dashed line), which indicates that most representations are biased toward classes with higher confidence and other positive classes with relatively low confidence are sidelined, causing inter-class imbalance and unreliable predictions. In the second column, we visualize the representations without confidence-based reweighting. Although the result shows balanced and distinctive representations, some prototypes are extremely close (circled in red dashed line). The reason is likely that the prototypes are calculated by naively averaging the representations and are disturbed by potential negative samples. Finally, our DCPGN dynamically assigns multiple pseudo labels to mitigate over-confident predictions and uses confidence-based reweighting to mitigate the interference of negative classes. In the third column, the class-wise representations and prototypes can be explicitly distinguished, which shows the effect of the proposed ML-PGM.

\begin{figure}[!t]
\centering
\includegraphics[width=1.0\linewidth]{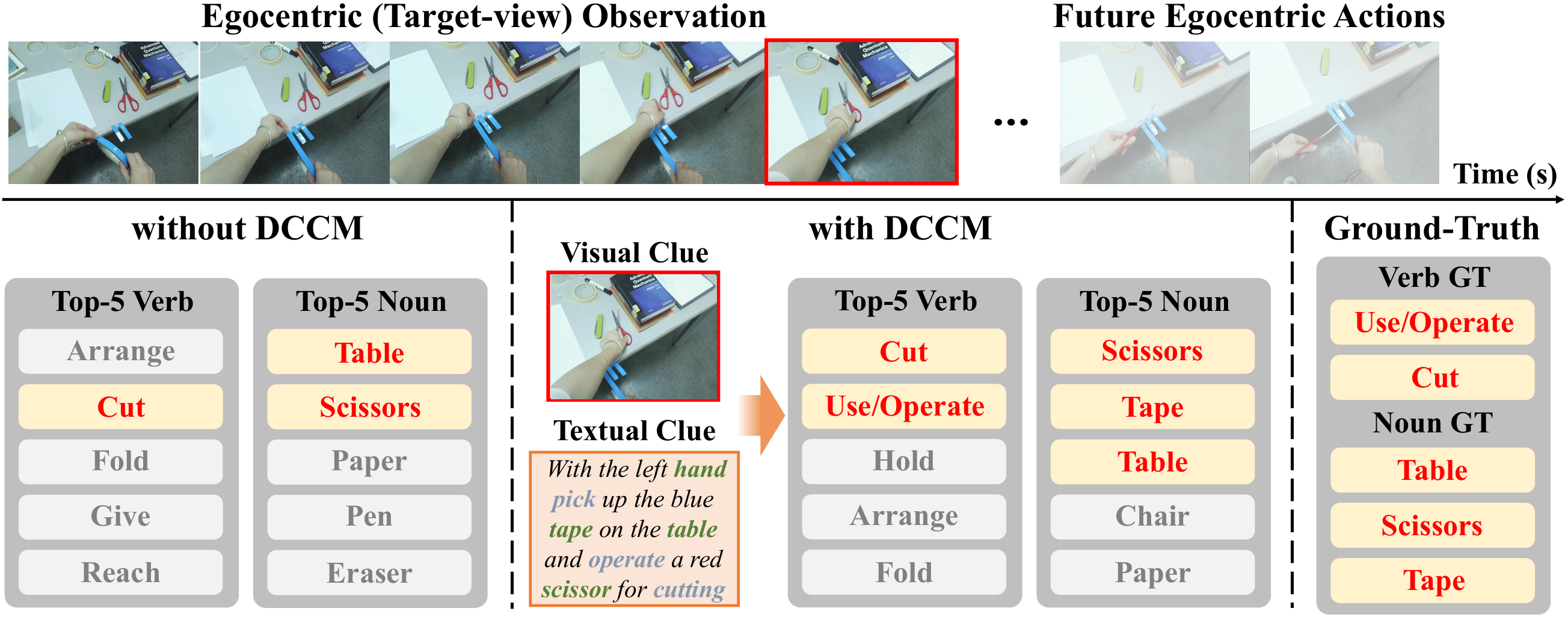}
\caption{Visualization of noun and verb anticipation candidates with or without DCCM under the Exo2Ego setting.}
\label{fig:4}
\end{figure}

\subsubsection{Analysis of Visual and Textual Clues}
Following the common evaluation setting \cite{huang2024egoexolearn}, the model is required to output five noun or verb candidates to anticipate the upcoming actions. To intuitively show the effectiveness of the proposed DCCM, we present an example to anticipate nouns and verbs under the Exo2Ego setting in Fig. \ref{fig:4}.

In detail, the target-view Ego observation shows the subject holding a blue tape and preparing to use a pair of red scissors, and the activity in the future is using the scissors on the table to cut the tape. Due to the significantly different object appearance and activity progress of different views, the model without DCCM only anticipates the verb ``cut" and nouns ``table", ``scissors" in the first column. In the second column, we introduce visual and textual clues that are rich in information about spatial objects and temporal action progressions. Then, we adopt an on-the-shelf CLIP to infer their respective logits and construct dual-clue consistency to bridge the inter-view gap. Therefore, the ground-truth nouns or verbs are accurately included in the predicted Top-5 candidates with the highest confidence scores among the candidates, demonstrating the effectiveness of our DCCM.

\section{Conclusion}
In this paper, we propose a novel TE$^{2}$A$^{3}$ task, which aims to adjust the source-view-trained model during testing to anticipate actions in the target view. To tackle this problem, we propose a Dual-Clue enhanced Prototype Growing Network (DCPGN), including a Multi-Label Prototype Growing Module (ML-PGM), which balances multiple positive classes via multi-label assignment and confidence-based reweighting under an entropy priority queue strategy. Then, we propose a Dual-Clue Consistency Module (DCCM) to construct dual-clue consistency by constraining the logits inferred from the textual and visual clues to bridge the Ego-Exo temporal-spatial view gap. Extensive experiments on the \textit{EgoExoLearn} and our newly proposed \textit{EgoMe-anti} demonstrate the effectiveness of our method, which outperforms other related works by a large margin.

\newpage

\section*{Acknowledgments}
This work was supported in part by the National Natural Science Foundation of China (No. U23A20286 and No. 62301121), Sichuan Science and Technology Program (No. 2026NSFSC1478), China Postdoctoral Science Foundation (No. 2025M783502 and No. GZB20240120).

\small
\bibliographystyle{ieeenat_fullname}
\bibliography{reference}

\normalsize 
\newpage

\maketitlesupplementary

\appendix
\setcounter{table}{0}   
\setcounter{figure}{0}
\setcounter{equation}{0}
\renewcommand{\thetable}{A\arabic{table}}
\renewcommand{\thefigure}{A\arabic{figure}}

\section{Details of State-of-the-Art Methods}
In this section, we introduce some technical and re-implementation details of the related state-of-the-art comparison methods.

\subsection{Tent}
Tent \cite{wang2020tent} is the pioneering work that utilizes target-domain data to adjust the trained model during test time via an entropy minimization strategy. It systematically defines the settings of fully test-time adaptation (TTA), i.e., fully TTA is independent of the training data and training loss, and adapts the model by optimizing an unsupervised loss during testing based on the unlabeled test data stream. Technically, it adjusts the affine parameters in batchnorm layers by minimizing the calculated entropy of the predicted logits. In this paper, we re-implement Tent for the TE$^{2}$A$^{3}$ task under the same settings as with our method. Due to the action anticipation model \cite{huang2024egoexolearn} not containing batchnorm layers, we practically select the parameters of the bias layers in the anticipation head, and the learning rate is set to 1e-4 for \textit{EgoExoLearn} and 1e-3 for \textit{EgoMe-anti} for optimization.

\subsection{TPT}
TPT \cite{shu2022test} is the first work to perform prompt tuning based on the CLIP model. TPT achieves test-time adaptation by augmenting a single test image to diverse views and minimizing entropy to optimize the learnable prompts. In this paper, we conduct re-implementation according to the ``TPT+CoOp" setting in this work. Since our TE$^{2}$A$^{3}$ task focuses on the adaptation of video rather than image, we adjust the original spatial cropping augmentation to the temporal re-sampling strategy to suit our task. In addition, we initialize the learnable prompts with ``a photo of a", whose length is 4 as in our work. For optimization, the learning rate is set to 1e-5 on the \textit{EgoExoLearn} and  1e-4 for the \textit{EgoMe-anti} benchmark.

\subsection{VITTA}
VITTA \cite{lin2023video} is the first TTA work for video-level action recognition. To address unanticipated distribution shifts of different-domain videos, VITTA first adopts a feature distribution alignment technique to align the estimated online statistics of the training and testing sets. Additionally, it constrains the predictions of different temporally augmented views to construct consistency. We re-implement this work according to its original settings. In detail, we first save the mean and variance for the features of the training set under all settings. Then, we employ momentum updating to obtain the mean and variance of the test set. Moreover, we adopt L1 losses to align the training and testing mean and variance statistics of each layer, and establish consistency between temporal augmented views. The learning rates are set to 1e-5 and 1e-4 for \textit{EgoExoLearn} and \textit{EgoMe-anti}, respectively. The quantitative results show that, even though VITTA accesses the statistics of the training set, our method still outperforms it by a large margin.

\subsection{TDA}
TDA \cite{karmanov2024efficient} is a training-free method for effective and efficient test-time adaptation. To tackle the distribution shifts between the source and target domains, TDA introduces a lightweight key-value cache and progressively refines the saved pseudo-labels for efficient adaptation to test data. Moreover, TDA also incorporates a negative cache branch to mitigate the impact of noise and achieves high performance for image-level TTA. In this paper, we adjust the hand-crafted prompt to be the same as our method (i.e., ``a photo of a") for a fair comparison. Furthermore, other hyperparameters such as threshold values for negative pseudo-labeling and testing feature selection are set following the official settings for re-implementation.

\subsection{ZERO}
ZERO \cite{farina2024frustratingly} is a simple yet effective method by setting the softmax temperature to zero. In detail, to address the model's poor generalization capability when presented with challenging examples, ZERO conducts thorough theoretical and empirical analysis. Technically, ZERO first augments the input image into multiple views, then feeds the images into the model to obtain anticipation logits. Next, it remains the confident predictions, and it finally sets the softmax temperature to zero. Following this paradigm, we conduct the temporal augmentation instead of the spatial augmentation specific to the video-level task. Then, we re-implement the ZERO method under its official settings for our TE$^{2}$A$^{3}$ task on the \textit{EgoExoLearn} and \textit{EgoMe-anti} benchmarks.  

\subsection{TCA}
TCA \cite{wang2024less} stands for Token Condensation Adaptation, which is an efficient training-free adaptation method to progressively refine the token selection process during test time. In detail, TCA first introduces a token reservoir to track and store class tokens sensitive to the domain shifts. Then, TCA dynamically adjusts multi-head attention maps based on the current data and accumulated knowledge. Finally, predictions are inferred not only from image-text similarity but also from correlations between visual features and tokens, which facilitates leveraging historical information for effective and efficient adaptation. In this paper, we re-implement this excellent work for the proposed TE$^{2}$A$^{3}$ task on the \textit{EgoExoLearn} and \textit{EgoMe-anti} benchmarks according to the official setting of TCA.

\subsection{ML-TTA}
ML-TTA \cite{wu2025multi} is a multi-label TTA method using the bound entropy minimization strategy. In detail, ML-TTA first describes the input images to determine the number of positive classes. Then, it infers a weak and a strong label set. Next, ML-TTA performs the label-binding strategy to bind the potential positive classes to mitigate the over-confident problem of plain entropy-based optimization, and conducts prompt learning for the two branches, respectively. The inferred logits of the two branches are added to obtain the final result. While effective, ML-TTA is for image-level TTA rather than video-level Ego-Exo adaptation and anticipation. In this paper, we directly utilize fine-level annotations from the original datasets \cite{huang2024egoexolearn,qiu2025egome} as descriptions for the input target-view videos. The learning rates are set to 1e-4 and 5e-4 for the \textit{EgoExoLearn} and \textit{EgoMe-anti} benchmarks for optimization. Despite our method employing the predicted descriptions while ML-TTA utilizes manually labeled caption annotations, our approach still surpasses ML-TTA by a large margin under all settings, which demonstrates the effectiveness of the proposed method.

\section{Details of Lightweight Narrator}
In this section, we introduce more architectural and implementation details of the lightweight narrator $\mathcal{N}(\cdot )$ in the Dual-Clue Consistency Module (DCCM) in Section 3.3.

The architecture of the $\mathcal{N}(\cdot)$ follows ``S2VTAttModel" in the official video-caption.pytorch library, which is a well-known open-source library for video captioning. The narrator is composed of an encoder and a decoder based on linear layers, GRU units, and attention layers. Assuming the input video frame features are denoted as $F^T$, we first convert $F^T$ into sequential features, as follows:
\begin{equation}
{{X}^{T}}=\{x_{1}^{T},x_{2}^{T},\cdots ,x_{{{L}}}^{T}\}={{\mathcal{F}}_{V}}({{F}^{T}})
\end{equation}
where ${{\mathcal{F}}_{V}}(\cdot )$ is an MLP comprising three linear layers, ${{X}^{T}}=\{x_{1}^{T},x_{2}^{T},\cdots ,x_{{{L}}}^{T}\}\in {{\mathbb{R}}^{{{L}}\times C}}$ means the converted input features, $L$ denotes the number of input video frames. Then, we utilize GRU units to model the temporal dependencies between the video frames. Taking the $t$-th timestep as an example, the calculation process is as follows:
\begin{equation}
{{r}_{t}}=s({{W}_{ir}}x_{t}^{T}+{{b}_{ir}}+{{W}_{hr}}{{\hat{h}}_{t-1}}+{{b}_{hr}})
\end{equation}
\begin{equation}
{{z}_{t}}=s({{W}_{iz}}x_{t}^{T}+{{b}_{iz}}+{{W}_{hz}}{{\hat{h}}_{t-1}}+{{b}_{hz}})
\end{equation}
\begin{equation}
{{n}_{t}}=\tanh({{W}_{in}}x_{t}^{T}+{{b}_{in}}+{{r}_{t}}*({{W}_{hn}}{{\hat{h}}_{t-1}}+{{b}_{hn}}))
\end{equation}
\begin{equation}
{{\hat{h}}_{t}}=(1-{{z}_{t}})*{{n}_{t}}+{{z}_{t}}*{{\hat{h}}_{t-1}}
\end{equation}
where $\hat{h}_t$ is the hidden state at the timestamp $t$, $s(\cdot )$ is the sigmoid function, $*$ is the Hadamard product, and $W$, $b$ are learnable weights and biases, respectively. The output of the encoder can be represented as $\hat{H}=\{{{\hat{h}}_{1}},{{\hat{h}}_{2}},\cdots {{\hat{h}}_{{{L}}}}\}\in {{\mathbb{R}}^{{{L}}\times {{C}_{h}}}}$, where $C_h$ is the dimension of the hidden state.

In the decoder, an attention mechanism is first applied:
\begin{equation}
{{O}_{1}}=\tanh ({{W}_{1}}[\hat{H};t({{\hat{h}}_{{{L}}}})]+{{b}_{1}})
\end{equation}
\begin{equation}
Att=\{{{\alpha }_{1}},{{\alpha }_{2}},\cdots {{\alpha }_{{{L}}}}\}=\sigma ({{W}_{2}}{{O}_{1}}+{{b}_{2}})
\end{equation}
\begin{equation}
c_{tx}=\sum\limits_{i=1}^{{{L}}}{{\alpha}_{i}\cdot \hat{h}_{i}}
\end{equation}
where $t(\cdot )$ is the tile operation, $\sigma$ is softmax function, $c_{tx} \in {{\mathbb{R}}^{{{C}_{h}}}}$ denotes the context feature. Then, we concatenate the context feature $c_{tx}$ with embeddings of the decoded words (each sentence begins with the $<$sos$>$ token), and initialize the hidden state of the decoder with the encoder's final hidden state ${{\hat{h}}_{{{L}}}}$. Finally, we adopt a GRU-based layer and a 2-layer MLP to decode the final sentence autoregressively.

For practical implementation, we collect view-agnostic video-text pairs $\{{{V}_{i}},{{T}_{i}}\}_{i=1}^{N}$ for training the narrator. In detail, we first remove all videos and corresponding annotations relevant to the data for test-time adaptation in the proposed TE$^{2}$A$^{3}$ task under all settings from the original EgoMe \cite{qiu2025egome} and EgoExoLearn \cite{huang2024egoexolearn} datasets. Then, we integrate the Ego and Exo video clips and the raw textual descriptions within the original dataset to construct the video captioning dataset $\{{{V}_{i}},{{T}_{i}}\}_{i=1}^{N}$. However, the description labeling rules are different for Ego and Exo videos in the EgoMe dataset (i.e., descriptions of Exo videos include attributes of the demonstrator). Therefore, we uniformly pre-process the texts in the video captioning dataset to concentrate on procedural activities and exclude potential view-specific information. Finally, we utilize the above video captioning dataset to train the lightweight narrator $\mathcal{N}(\cdot )$ to generate view-agnostic textual clues, facilitating the adaptation between Ego and Exo views during test time. Specifically, we set the feature dimension $C_h$ to 512. The maximum sentence length is set to 28, and each sentence begins with the $<$sos$>$ token and ends with the $<$eos$>$ token.  Additionally, the sampling strategy of video frame features is identical to that in the action anticipation network (i.e., uniformly sample 5 frames as input). We train the narrator network for 1000 epochs with a batch size of 1024. The dropout rate is 0.5 and 0.2 for the linear and GRU-based layers, respectively, which helps alleviate overfitting. Moreover, we set the initial learning rate to 5e-4 and apply a learning rate decay strategy, in which the learning rate is reduced to 80$\%$ of its original value every 200 epochs.

\begin{table*}[!t]
\centering
\caption{The number of samples under each view of the \textit{train}/\textit{val}/\textit{test} set of the \textit{EgoMe-anti} and \textit{EgoExoLearn} benchmarks.}
\label{tab:a1}
\scalebox{1.0}{
\begin{tabular}{p{2.5cm}<{\raggedright}|p{1.2cm}<{\centering}p{1.2cm}<{\centering}|p{1.2cm}<{\centering}p{1.2cm}<{\centering}|p{1.2cm}<{\centering}p{1.2cm}<{\centering}|p{1.2cm}<{\centering}p{1.2cm}<{\centering}}
\toprule
\multirow{3}{*}{Subset} & \multicolumn{4}{c|}{\textit{EgoMe-anti}}  & \multicolumn{4}{c}{\textit{EgoExoLearn}}                       \\ 
     & \multicolumn{2}{c}{Exo} & \multicolumn{2}{c|}{Ego}   & \multicolumn{2}{c}{Exo}     & \multicolumn{2}{c}{Ego}     \\ \cmidrule{2-9} 
     & Noun & Verb & Noun & Verb & Noun & Verb & Noun & Verb \\ \midrule
     Train set & 8336 & 8336 & 8939  & 8939 & 11762	& 17516 & 78055 & 33496 \\       
     Validation set & 1817 & 1817 & 1970  & 1970 & 8235	& 1960 & 84726 & 91454 \\    
     Test set & 3584 & 3584 & 3889  & 3889 & 12395	& 5371 & 15231 & 50471 \\       
\bottomrule
\end{tabular}}
\end{table*}

\section{Details of the Benchmarks}
In this section, we show more details of our newly proposed \textit{EgoMe-anti} and the existing \textit{EgoExoLearn} benchmarks to support the TE$^{2}$A$^{3}$ task.

\subsection{EgoMe-anti}
The \textit{EgoMe-anti} benchmark is constructed based on the recent EgoMe \cite{qiu2025egome} dataset, which follows the imitation learning process in real-world scenarios. In detail, it contains 82.8 hours of Exo observation and Ego following videos and covers up to 41 real-world scenarios such as the library, classroom, gym, and so on. The video durations range widely from 3 seconds to over 60 seconds, and most of the videos last from 10 to 25 seconds. Moreover, each video contains 2 to above 15 fine-level consecutive and non-overlapped atomic actions with timestamps and descriptions. We exclude the incorrectly following videos and leverage the fine-level timestamps and description annotations to construct the \textit{EgoMe-anti} benchmark, whose sample numbers of the \textit{train}/\textit{val}/\textit{test} sets are shown in the left panel of Table \ref{tab:a1}. Because we extract nouns and verbs from the raw sentences simultaneously, the number of samples remains identical in noun or verb anticipation tasks under the same perspective. Finally, we summarize 35 noun and 29 verb categories for the action anticipation task.

\subsection{EgoExoLearn}
The \textit{EgoExoLearn} benchmark in our work directly follows the official cross-view action anticipation benchmark proposed in EgoExoLearn \cite{huang2024egoexolearn}, which is an excellent dataset containing large-scale asynchronous Ego and Exo videos of procedural activities in daily and professional scenarios. Specifically, it comprises up to 745 long videos with a total duration of around 120 hours. The video recording is oriented towards 5 types of daily cooking tasks and 3 types of chemical laboratory tasks, which are recorded in 4 kinds of different kitchens and 3 kinds of laboratories to ensure data diversity. This dataset also contains detailed coarse-grained and fine-grained timestamps, descriptions, and noun/verb categories. Therefore, the \textit{EgoExoLearn} benchmark for the action anticipation task is constructed based on the fine-grained annotations with massive samples, and the number of samples of each subset in the Ego or Exo perspectives is represented in the right panel in Table \ref{tab:a1}. In addition, \textit{EgoExoLearn} has 31 noun and 19 verb categories for the action anticipation task.

\begin{table*}[!t]
\centering
\caption{Quantitative results on the \textit{EgoMe-anti} and \textit{EgoExoLearn} under the Exo2Ego and Ego2Exo settings in Top-1 recall evaluation.}
\label{tab:a1_cr}
\scalebox{0.95}{
\begin{tabular}{p{3.7cm}<{\raggedright}|p{1.2cm}<{\centering}p{1.2cm}<{\centering}|p{1.2cm}<{\centering}p{1.2cm}<{\centering}|p{1.2cm}<{\centering}p{1.2cm}<{\centering}|p{1.2cm}<{\centering}p{1.2cm}<{\centering}}
\toprule
\multirow{3}{*}{Methods} & \multicolumn{4}{c|}{\textit{EgoMe-anti}}  & \multicolumn{4}{c}{\textit{EgoExoLearn}}                                                         \\ 
     & \multicolumn{2}{c}{Exo2Ego} & \multicolumn{2}{c|}{Ego2Exo}   & \multicolumn{2}{c}{Exo2Ego}     & \multicolumn{2}{c}{Ego2Exo}     \\ \cmidrule{2-9} 
     & Noun & Verb & Noun & Verb & Noun & Verb & Noun & Verb \\ \midrule
     \rowcolor{gray!15} Ours without Adaptation & 27.30 & 6.24 & 22.72  & 6.45 & 6.07	&  5.43& 6.90& 5.35 \\
     Tent \cite{wang2020tent}  & 31.72 & 8.09 & 26.92  & 8.98 & 7.72	& 6.57 & 7.98 & 5.40 \\
     TPT \cite{shu2022test} & 31.97 & 8.05 &  27.14 & 8.89 & 8.11	& 6.35 & 8.13 & 5.43 \\
     VITTA \cite{lin2023video}  & 32.21 & 8.75 &  26.87 & 9.14 & 7.49	& 6.13 & 8.27 & 5.56 \\
     TDA \cite{karmanov2024efficient} & 33.42 & \underline{10.60} & 27.18  &  \underline{10.46} & 8.26	& 6.59 & 8.41 & 5.48 \\
     ZERO \cite{farina2024frustratingly}  & 32.50 & 10.02 &  25.82 & 10.26 & 9.48&  \underline{8.73} & \underline{10.27} & \underline{7.42}\\
     TCA \cite{wang2024less}  & 32.35 & 8.45 & 27.36  & 9.07 & 7.85	& 6.22 & 8.05 & 5.55 \\
     ML-TTA \cite{wu2025multi}  &  \underline{33.48} & 10.37 & \underline{27.80}  & 10.06 & \underline{10.39}	& 7.40 & 10.01 &5.38 \\
     \textbf{DCPGN (Ours)} & \textbf{38.00} & \textbf{20.48} & \textbf{33.23}  &  \textbf{17.75} & \textbf{17.14}	&  \textbf{11.55} & \textbf{16.30} &  \textbf{14.23} \\            
\bottomrule
\end{tabular}}
\end{table*}

\begin{table*}[!t]
\centering
\caption{Analysis of the maximum capacity $N$ of the memory bank. Note that the saved data in the memory banks is also included in the statistics of the number of parameters.}
\label{tab:a2}
\scalebox{1.0}{
\begin{tabular}{p{1.0cm}<{\centering}|p{1.2cm}<{\centering}p{1.2cm}<{\centering}|p{1.2cm}<{\centering}p{1.2cm}<{\centering}|p{1.2cm}<{\centering}p{1.2cm}<{\centering}|p{1.2cm}<{\centering}p{1.2cm}<{\centering}|p{2.0cm}<{\centering}}
\toprule
\multirow{3}{*}{$N$} &  \multicolumn{4}{c|}{\textit{EgoMe-anti}}  & \multicolumn{4}{c|}{\textit{EgoExoLearn}}  & \multirow{3}{*}{$\#$Params (M)}                                                   \\ 
     &  \multicolumn{2}{c}{Exo2Ego} & \multicolumn{2}{c|}{Ego2Exo}   & \multicolumn{2}{c}{Exo2Ego}     & \multicolumn{2}{c|}{Ego2Exo}     \\ \cmidrule{2-9} 
     &  Noun & Verb & Noun & Verb & Noun & Verb & Noun & Verb \\ \midrule
    100  & 78.99  &	\underline{43.64} & \underline{71.97}	& 39.62 & 45.07 & 42.60  &  47.48  &	45.58 & 1.71 \\ 
    200  & \textbf{79.24}  &	42.47  & 71.97	& 39.71  &45.36  & 42.72  &  47.87  &	45.61 & 3.42 \\ 
    300  & 78.95  &	42.95 &  71.81	& \underline{39.82} & 45.50 & \textbf{43.00}  &  \underline{48.18}  &  45.99	 & 5.13 \\ 
    400  & 78.86  &	43.17 &  71.57	& 38.95  & \underline{45.78} &  42.74 &  48.06  &	\underline{46.10}  & 6.84 \\ 
    500  &\underline{79.03}&	\textbf{43.84 }&\textbf{72.01}	&\textbf{40.10} &\textbf{46.26}	&\underline{42.98} & \textbf{48.48}&	\textbf{46.51} & 8.54 \\         
\bottomrule
\end{tabular}}
\end{table*}

\section{Results in Top-1 Recall Evaluation}
In addition to the common Top-5 recall evaluation metric \cite{huang2024egoexolearn,damen2022rescaling}, we also evaluate our DCPGN and comparison methods under the Top-1 recall metric for more comprehensive evaluation. As shown in Table \ref{tab:a1_cr}, our method still outperforms other comparison methods by a large margin. In detail, it surpasses the most recent ML-TTA \cite{wu2025multi} by 4.52$\%$, 10.11$\%$, 5.43$\%$, 7.69$\%$ on the \textit{EgoMe-anti} benchmark, and by 6.75$\%$, 4.15$\%$, 6.29$\%$, 8.85$\%$ on the \textit{EgoExoLearn} benchmark, which further demonstrates the effectiveness and superiority of the proposed DCPGN.

\section{Analysis of Hyperparameters}
In this section, we conduct comprehensive experiments to analyze several hyperparameters in the proposed method. Moreover, we report the results under Exo2Ego and Ego2Exo settings on the \textit{EgoMe-anti} and \textit{EgoExoLearn} benchmarks.

\subsection{Analysis of memory bank capacity $N$}
In Table \ref{tab:a2}, we conduct experiments under different settings of the hyperparameter $N$, which denotes the maximum capacity of the memory bank for each class. Specifically, we set the value of $N$ to 100, 200, 300, 400, and 500 and report the number of parameters (including the saved data in the memory banks) and quantitative results under all settings on the \textit{EgoMe-anti} and \textit{EgoExoLearn} benchmarks. The experimental results show that the performances fluctuate within a relatively small range, and the number of parameters is tolerable under all settings of $N$. It demonstrates the effectiveness of the proposed memory banks in the ML-PGM, which can achieve high performance even though the memory bank capacity is limited. Moreover, the model reaches the highest performance when $N$ is 500. Considering its superior performance and tolerable number of parameters, we finally set $N$ to 500 in our DCPGN.

\subsection{Analysis of hyperparameter $\alpha$}
In Table \ref{tab:a3}, we conduct quantitative experiments under all settings on the two benchmarks to analyze the hyperparameter $\alpha$, which is used to balance the distinct logits from ML-PGM and DCCM. Keeping other hyperparameters and settings unchanged, we set the balance coefficient $\alpha$ to 0.25, 0.50, 0.75, and 1.00, and report their respective results. The results in Table \ref{tab:a3} show that the model achieves the best performance when $\alpha$ is set to 0.5. Furthermore, the model is not sensitive to the settings of $\alpha$ within the range from 0.25 to 1.0, which indicates the robustness of the model to different settings of the balancing hyperparameter, demonstrating the effectiveness of the proposed method.

\begin{table*}[!t]
\centering
\caption{Analysis of the hyperparameter $\alpha$ for balancing distinct logits inferred by ML-PGM and DCCM.}
\label{tab:a3}
\scalebox{1.0}{
\begin{tabular}{p{1.2cm}<{\centering}|p{1.2cm}<{\centering}p{1.2cm}<{\centering}|p{1.2cm}<{\centering}p{1.2cm}<{\centering}|p{1.2cm}<{\centering}p{1.2cm}<{\centering}|p{1.2cm}<{\centering}p{1.2cm}<{\centering}}
\toprule
\multirow{3}{*}{$\alpha$} & \multicolumn{4}{c|}{\textit{EgoMe-anti}}  & \multicolumn{4}{c}{\textit{EgoExoLearn}}                                                         \\ 
     & \multicolumn{2}{c}{Exo2Ego} & \multicolumn{2}{c|}{Ego2Exo}   & \multicolumn{2}{c}{Exo2Ego}     & \multicolumn{2}{c}{Ego2Exo}     \\ \cmidrule{2-9} 
     & Noun & Verb & Noun & Verb & Noun & Verb & Noun & Verb \\ \midrule
     0.25 & 78.61 & \underline{42.64} & \underline{71.85} &  \underline{39.21} &44.44 & \underline{42.89} & 47.70  &	46.22 \\ 
     0.50 & \underline{79.03}&	\textbf{43.84 }&\textbf{72.01}	&\textbf{40.10} &\textbf{46.26}	&\textbf{42.98} & \textbf{48.48}&	\textbf{46.51} \\ 
     0.75 & \textbf{79.19}  & 41.62 & 71.09 & 37.81 & \underline{45.39} 	& 42.62 & \underline{48.43}  &	\underline{46.31} \\ 
     1.00 &  78.17& 40.89 & 71.37 & 38.31 & 45.37	& 41.98 & 48.13 & 46.11	 \\ 
\bottomrule
\end{tabular}}
\end{table*}

\begin{table*}[!t]
\centering
\caption{Analysis of the hyperparameter $\mu_1$ for scaling the visual logits in the DCCM.}
\label{tab:a4}
\scalebox{1.0}{
\begin{tabular}{p{1.2cm}<{\centering}|p{1.2cm}<{\centering}p{1.2cm}<{\centering}|p{1.2cm}<{\centering}p{1.2cm}<{\centering}|p{1.2cm}<{\centering}p{1.2cm}<{\centering}|p{1.2cm}<{\centering}p{1.2cm}<{\centering}}
\toprule
\multirow{3}{*}{$\mu_1$} & \multicolumn{4}{c|}{\textit{EgoMe-anti}}  & \multicolumn{4}{c}{\textit{EgoExoLearn}}                                                         \\ 
     & \multicolumn{2}{c}{Exo2Ego} & \multicolumn{2}{c|}{Ego2Exo}   & \multicolumn{2}{c}{Exo2Ego}     & \multicolumn{2}{c}{Ego2Exo}     \\ \cmidrule{2-9} 
     & Noun & Verb & Noun & Verb & Noun & Verb & Noun & Verb \\ \midrule
     0.25 & 77.52 & \underline{43.16} & 71.41 & 38.90 & 45.36 & 42.82 & 48.10  &	\underline{46.26} \\ 
     0.50 & \underline{78.57} & 42.62 & 71.67 & \underline{39.62} & 45.53 & \underline{42.97} & 48.08  & 45.89    \\ 
     0.75 & 78.47 & 42.37 & \underline{71.82} & 39.40 & \underline{45.59} & 42.93 & \underline{48.38}  &	45.66 \\ 
     1.00 & \textbf{79.03}&	\textbf{43.84 }&\textbf{72.01}	&\textbf{40.10} &\textbf{46.26}	&\textbf{42.98} & \textbf{48.48}&	\textbf{46.51}	 \\ 
\bottomrule
\end{tabular}}
\end{table*}

\begin{table*}[!t]
\centering
\caption{Analysis of the hyperparameter $\mu_2$ for scaling the textual logits in the DCCM.}
\label{tab:a5}
\scalebox{1.0}{
\begin{tabular}{p{1.2cm}<{\centering}|p{1.2cm}<{\centering}p{1.2cm}<{\centering}|p{1.2cm}<{\centering}p{1.2cm}<{\centering}|p{1.2cm}<{\centering}p{1.2cm}<{\centering}|p{1.2cm}<{\centering}p{1.2cm}<{\centering}}
\toprule
\multirow{3}{*}{$\mu_2$} & \multicolumn{4}{c|}{\textit{EgoMe-anti}}  & \multicolumn{4}{c}{\textit{EgoExoLearn}}                                                         \\ 
     & \multicolumn{2}{c}{Exo2Ego} & \multicolumn{2}{c|}{Ego2Exo}   & \multicolumn{2}{c}{Exo2Ego}     & \multicolumn{2}{c}{Ego2Exo}     \\ \cmidrule{2-9} 
     & Noun & Verb & Noun & Verb & Noun & Verb & Noun & Verb \\ \midrule
     0.25 & 78.48 & \underline{42.98}  & \underline{71.90} & \underline{39.37} & 43.99 & 42.62 &  47.90  &	46.14 \\ 
     0.50 & \textbf{79.03}&	\textbf{43.84 }&\textbf{72.01}	&\textbf{40.10} &\underline{46.26}	&\underline{42.98} & \textbf{48.48}&	\textbf{46.51}	 \\ 
     0.75 & \underline{78.86} & 42.76 & 71.87 & 39.05 & \textbf{46.27}  & 42.78 & 48.34  & \underline{46.14}	 \\ 
     1.00 & 77.99 & 42.39 & 70.60 & 39.21 & 44.33 & \textbf{43.01}  & \underline{48.39} &	45.68 \\ 
\bottomrule
\end{tabular}}
\end{table*}

\subsection{Analysis of hyperparameter $\mu_1$}
In Table \ref{tab:a4}, we quantitatively analyze the hyperparameter $\mu_1$, which is used to scale the visual logits inferred based on the visual clue in the proposed DCCM. In detail, we conduct experiments and report the performances under Exo2Ego and Ego2Exo settings on the two benchmarks, where the hyperparameter $\mu_1$ is set to 0.25, 0.5, 0.75, and 1.0, respectively, and other hyperparameters remain unchanged. The results show that the model exhibits robustness to the change of $\mu_1$, i.e., the metrics fluctuate within 1.5$\%$. Finally, we set $\mu_1$ to 1.0, in which case the model yields optimal performance under both view adaptation settings.

\subsection{Analysis of hyperparameter $\mu_2$}
In Table \ref{tab:a5}, we set the hyperparameter $\mu_2$ for scaling the textual logits to 0.25, 0.5, 0.75, and 1.0, and conduct experiments to analyze its effects comprehensively. The experimental results show that when the hyperparameter $\mu_2$ is set to 0.5, the model achieves the best performance under all settings on \textit{EgoMe-anti}. For the \textit{EgoExoLearn} benchmark, the model yields the best metrics under the Ego2Exo setting and second-place results under the Exo2Ego setting. Therefore, we finally assign the hyperparameter $\mu_2$ to 0.5 in our model. In addition, the final performance remains stable when the hyperparameter $\mu_2$ is changed from 0.25 to 1.0, which demonstrates the robustness of the proposed DCPGN to different $\mu_2$ values.

\subsection{Analysis of the number of test samples $N_{test}$}
The number of test samples $N_{test}$ is crucial for the memory banks in the Multi-Label Prototype Growing Module (ML-PGM) to create reliable and meaningful prototypes. Therefore, it is necessary to ablate the number of samples during the test-time adaptation process. In Table \ref{tab:a7}, we conduct ablation experiments of the number of test samples $N_{test}$ on the \textit{EgoMe-anti} benchmark with a fixed random seed, which show that reliable prototypes can be created when using only 25$\%$ samples of the full test set (972 Ego or 896 Exo samples), with only a slight performance drop compared to using the full test set. It further demonstrates the effectiveness of the proposed ML-PGM, which can achieve promising performance even with a small test set.

\begin{table}[htbp]
\centering
\caption{Ablation on the number of test samples on \textit{EgoMe-anti} benchmark (with 3889 Ego and 3584 Exo test samples in total).}
\label{tab:a7}
\scalebox{1.0}{
\begin{tabular}{p{1.5cm}<{\raggedright}|p{1.0cm}<{\centering}p{1.0cm}<{\centering}|p{1.0cm}<{\centering}p{1.0cm}<{\centering}}
\toprule
\multirow{2}{*}{$N_{test}$} 
 & \multicolumn{2}{c|}{Exo2Ego}     & \multicolumn{2}{c}{Ego2Exo}     \\ \cmidrule{2-5} 
 & Noun & Verb & Noun & Verb \\ \midrule
Full  & \textbf{79.03}	& \textbf{43.84} & \textbf{72.01}   & \textbf{40.10} \\
50$\%$  & 78.59	& 42.76 &  71.38  &  39.28\\
25$\%$  & 78.41	& 42.38 &  71.01 &  38.76\\
10$\%$  & 75.42	& 38.32 &  64.84  &  35.75\\
5$\%$  & 72.84	& 36.08 &  63.40  &  33.51\\ 
\bottomrule
\end{tabular}}
\end{table}

\section{More Visualization Results}
In this section, we show more visualization results, such as the t-SNE visualization results and the noun/verb anticipation examples to verify the effectiveness of our method.

\subsection{t-SNE visualizations}
\begin{figure*}[!t]
\centering
\includegraphics[width=1.0\linewidth]{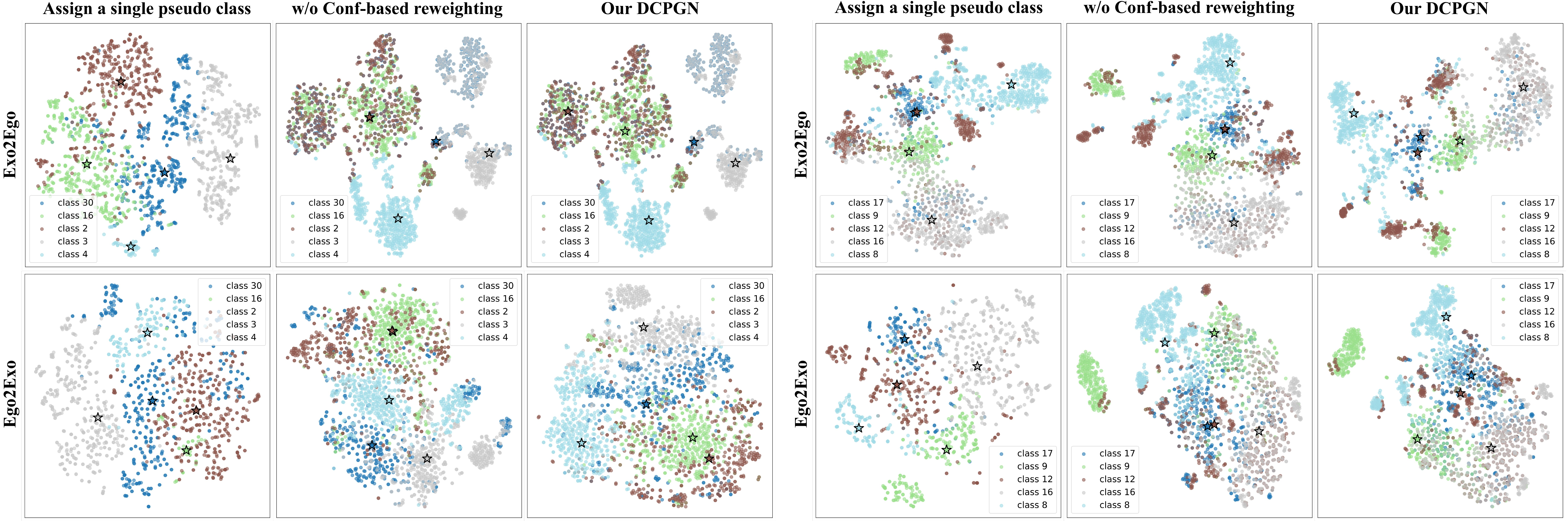}
\caption{t-SNE visualization of representations in the memory banks and the corresponding prototypes of five dominant classes under Exo2Ego and Ego2Exo settings. The left panel shows the results on the \textit{EgoExoLearn} benchmark and the right panel shows the results on the \textit{EgoMe-anti} benchmark. Dots denote class-wise representations and stars denote prototypes (best viewed in color).}
\label{fig:a1}
\end{figure*}

We conduct more experiments by means of t-SNE \cite{van2008visualizing} to visualize representations in the memory banks and the corresponding prototypes under all settings on the \textit{EgoExoLearn} and \textit{EgoMe-anti} benchmarks in Fig. \ref{fig:a1}. Practically, we visualize five dominant classes with the most samples following the settings in Section 4.4.3 for ease of analysis.

We visualize representations and prototypes of the model that only assigns a single class, the model without confidence-based reweighting, and the final DCPGN model. In the first column, the visualization results show that some classes are assigned only a small number of representations, and prototypes of different classes are highly overlapping in some cases. It further verifies that the single-label assignment strategy may lead to remarkable inter-class imbalance and unreliable prototypes, which cause performance degradation. The results of the model without confidence-based reweighting are presented in the second column, which show that distinct classes form balanced representations and can be well distinguished. Note that it is normal for overlapping representation clusters of different classes, because each selected representation sample is assigned to multiple pseudo labels. However, due to the interference of the potential negative samples and highly overlapping representations across many classes, some prototypes are extremely close, which impairs the model's class-wise discriminative capability. Finally, in the third column, we visualize the representations and prototypes of our DCPGN. The results show that the representations are balanced and prototypes can be explicitly distinguished, which demonstrates the effectiveness of the multi-label assignment and confidence-based reweighting strategies in the ML-PGM of the proposed DCPGN.

\subsection{Noun/verb anticipation visualizations}

\begin{figure*}[!t]
\centering
\includegraphics[width=1.0\linewidth]{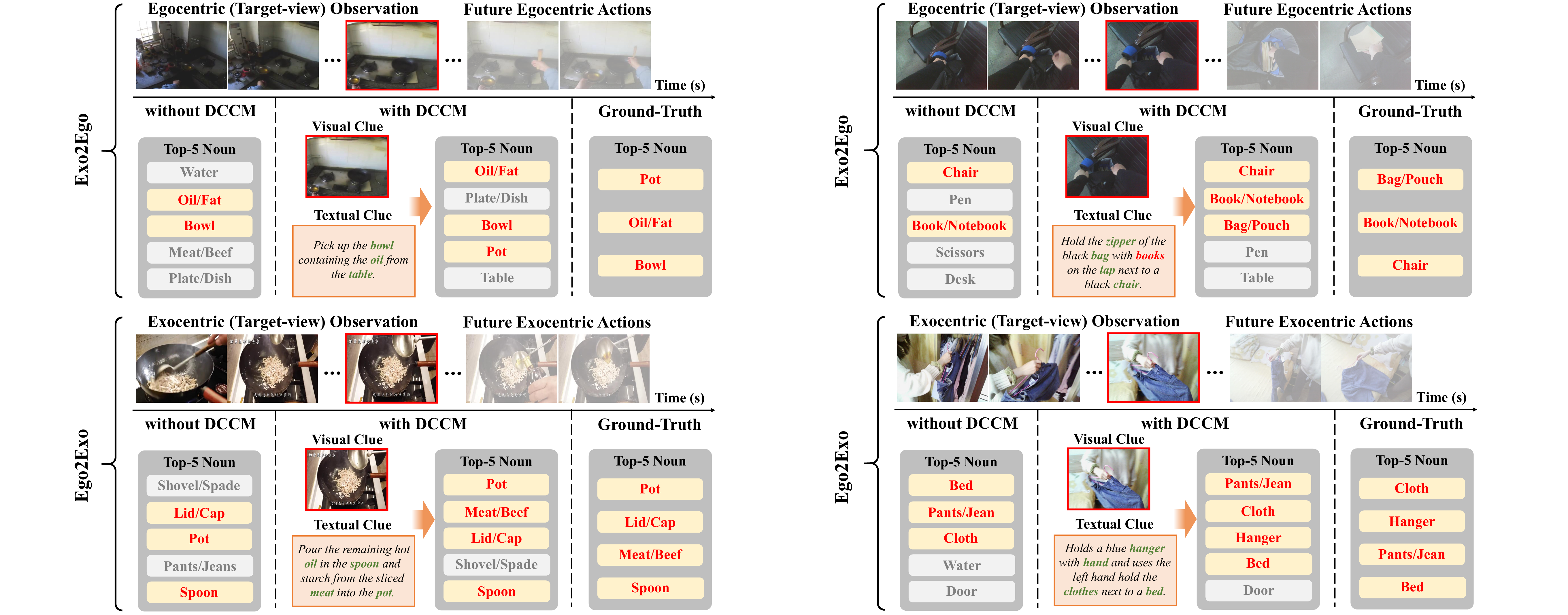}
\caption{Visualization of the \textit{Top-5} candidates for \textbf{noun anticipation} with or without the proposed DCCM under the Exo2Ego and Ego2Exo setting. The left panel shows the results on the \textit{EgoExoLearn} benchmark, while the right panel shows the results on \textit{EgoMe-anti}.}
\label{fig:a2_1}
\end{figure*}

\begin{figure*}[!t]
\centering
\includegraphics[width=1.0\linewidth]{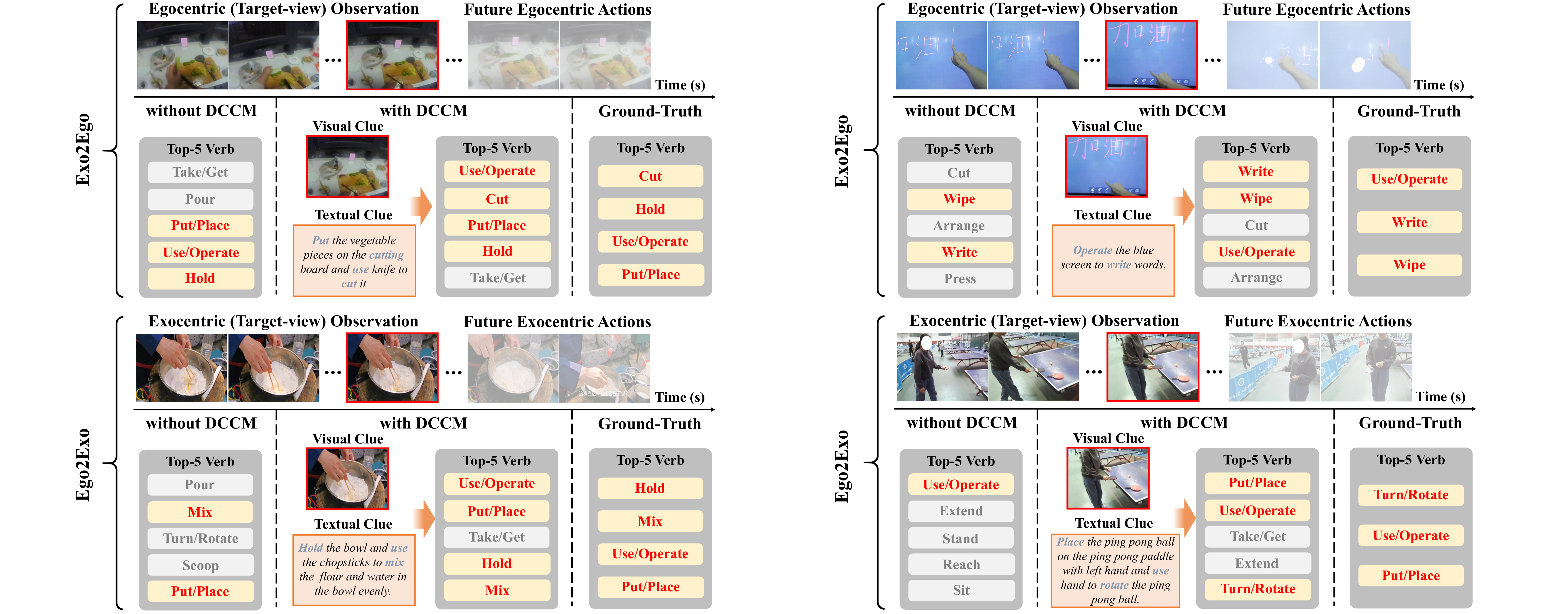}
\caption{Visualization of the \textit{Top-5} candidates for \textbf{verb anticipation} with or without the proposed DCCM under the Exo2Ego and Ego2Exo setting. The left panel shows the results on the \textit{EgoExoLearn} benchmark, while the right panel shows the results on \textit{EgoMe-anti}.}
\label{fig:a2_2}
\end{figure*}

In this section, we conduct more experiments to visualize the noun and verb anticipation results of the model with or without DCCM in Fig. \ref{fig:a2_1} and Fig. \ref{fig:a2_2}, respectively. We visualize the \textit{Top-5} noun/verb anticipation candidates and the corresponding visual and textual clues under the Exo2Ego and Ego2Exo settings on \textit{EgoExoLearn} and \textit{EgoMe-anti}.

\subsubsection{Noun anticipation results}
In Fig. \ref{fig:a2_1}, we comprehensively visualize the results of the anticipated noun candidates and the corresponding visual and textual clues in the proposed DCCM. The experimental results show that visual and textual clues can effectively convey information about nouns in the upcoming actions under both Exo2Ego and Ego2Exo settings on the two benchmarks. In the example under the Exo2Ego setting on the \textit{EgoExoLearn} benchmark (left panel), the model without DCCM cannot anticipate the active noun ``pot" in the future. However, the visual clue in DCCM clearly contains the ``pot", which facilitates the subsequent noun anticipation. For the Ego2Exo example on \textit{EgoExoLearn}, due to the visual clue clearly presenting ``meat" in the pot and the textual clue comprising the word ``meat", the model with DCCM can precisely anticipate the ``meat/beef" class, which is active in the following stir-frying process. For the examples on the \textit{EgoMe-anti} benchmark (right panel), in the upper Exo2Exo example, the visual and textual clues of DCCM both present the ``book" semantic, and successfully anticipate the active ``book/notebook" class. Similarly, in the lower Ego2Exo example, the model without DCCM fails to anticipate the ``hanger" class, which is an important object in the cloth-picking process. Nevertheless, the DCCM provides clues of ``hanger" in both visual and textual modalities and accurately anticipates the corresponding noun class. The above visualization examples demonstrate that the proposed DCCM can explicitly mitigate the spatial gap between the Ego and Exo perspectives, thereby facilitating accurate noun anticipation.

\subsubsection{Verb anticipation results}
In Fig. \ref{fig:a2_2}, we conduct more experiments to visualize the anticipated verbs as well as the visual and textual clues under both Exo2Ego and Ego2Exo settings on the \textit{EgoExoLearn} (left panel) and \textit{EgoMe-anti} (right panel) benchmarks. Because it is difficult for single-frame visual clues to reflect the temporal activity progressions, textual clues become more essential in the verb anticipation task. In the example under the Exo2Ego setting on \textit{EgoExoLearn}, the model without DCCM fails to anticipate the verb ``cut", whereas the textual clues in the DCCM clearly describe the process of cutting the vegetable, which leads to accurate anticipation of this verb class. In the Ego2Exo example on \textit{EgoExoLearn}, the textual clue includes the verbs ``hold" and ``use", which facilitate more accurate anticipated verb candidates compared to the model without DCCM. The right panel presents examples of the \textit{EgoMe-anti} benchmark. Specifically, the upper Exo2Ego example shows the subject operating the screen to write and wipe words, and the textual clue contains words such as ``operate" and ``write", which are consistent with classes of the upcoming actions. Finally, the lower Ego2Exo example shows that a girl will place the ball in her hand and then use the paddle to turn the ball. The results demonstrate that our model with DCCM yields more accurate verb anticipation candidates than those output by the model without DCCM. The above cases further demonstrate the effectiveness of DCCM in alleviating the temporal gap between distinct views, leading to more accurate anticipation for verb classes of the upcoming fine-level actions.

\newpage

\end{document}